\documentclass[sigconf]{acmart}
\AtBeginDocument{%
  }
\usepackage{enumitem}
\usepackage{makecell}
\usepackage{multirow}
\usepackage[table]{xcolor}
\usepackage{booktabs}
\usepackage{tabularx}
\usepackage{graphicx}
\definecolor{softblue}{RGB}{230, 242, 255}
\newcounter{protocol} 
\renewcommand{\theprotocol}{\arabic{protocol}} 
\newcolumntype{Y}{>{\centering\arraybackslash}X}
\newcolumntype{H}{>{\hsize=1.9\hsize\centering\arraybackslash}X}
\newcolumntype{S}{>{\hsize=0.85\hsize\centering\arraybackslash}X}
\copyrightyear{2026}
\acmYear{2026}
\setcopyright{cc}
\setcctype{by}
\acmConference[MM '26]{Proceedings of the 34th ACM International Conference on Multimedia}{November 10--14, 2026}{Rio de Janeiro, Brazil}
\acmBooktitle{Proceedings of the 34th ACM International Conference on Multimedia (MM '26), November 10--14, 2026, Rio de Janeiro, Brazil}
\acmDOI{10.1145/3767308.3835218}
\acmISBN{979-8-4007-2213-4/2026/11}

\begin{document}

\title{SphereVideo: Prototype-anchored Hyperspherical Boundary for Continual AI-generated Video Detection}

\author{Fei Li}
\orcid{0009-0000-8530-1985}
\affiliation{%
  \institution{Fudan University}
  \city{Shanghai}
  \country{China}
}
\email{lif25@m.fudan.edu.cn}

\author{Yue Yu}
\orcid{0009-0003-1416-783X}
\affiliation{%
  \institution{Fudan University}
  \city{Shanghai}
  \country{China}
}
\email{yuy24@m.fudan.edu.cn}

\author{Yuran Wang}
\orcid{0000-0002-3065-2830}
\affiliation{%
  \institution{Peking University}
  \city{Beijing}
  \country{China}
}
\email{yuranwang25@stu.pku.edu.cn }

\author{Xinghan Li}
\orcid{0009-0004-9785-632X}
\affiliation{%
  \institution{Fudan University}
  \city{Shanghai}
  \country{China}
}
\email{xinghanli24@m.fudan.edu.cn}

\author{Jingjing Chen}
\orcid{0000-0003-3148-264X}  
\authornote{Corresponding author.} 
\affiliation{%
  \institution{Fudan University}
  \city{Shanghai}
  \country{China}
}
\email{chenjingjing@fudan.edu.cn}

\author{Yu-Gang Jiang}
\orcid{0000-0002-1907-8567}
\affiliation{%
  \institution{Fudan University}
  \city{Shanghai}
  \country{China}
}
\email{ygj@fudan.edu.cn}


\begin{abstract}
AI-generated video (AIGV) detection aims to distinguish real videos from AI-generated ones. In practice, detectors trained on existing data often fail to generalize to newly emerging generative models, making this task challenging. Therefore, continual learning (CL) is essential for improving the adaptability. However, CL frameworks for this task remain underexplored. To this end, we propose SphereVideo, a novel CL framework for AIGV detection built on two key observations. First, real videos exhibit a compact feature distribution. Based on this, we encourage real video features to cluster around a real prototype on a hypersphere while repelling AI-generated samples, thereby establishing a decision boundary. This prototype serves as a stable anchor for CL, regulating boundary evolution and mitigating catastrophic forgetting. Second, existing methods tend to rely solely on spatial artifacts as shortcuts. To enhance temporal modeling, we introduce a strategy that models the temporal dynamics of real data at both frame and clip levels. By strengthening real data modeling, this strategy further facilitates learning a real prototype and forming a stable decision boundary. Moreover, we construct a comprehensive and challenging benchmark. 
Extensive experiments demonstrate that SphereVideo achieves an improved plasticity-stability trade-off, outperforming prior methods by 3.08\% on seen data and 4.00\% on unseen AI-generated data.
\end{abstract}



\maketitle
\section{Introduction}
\begin{figure}[t]  
  \centering
  \includegraphics[width=\linewidth]{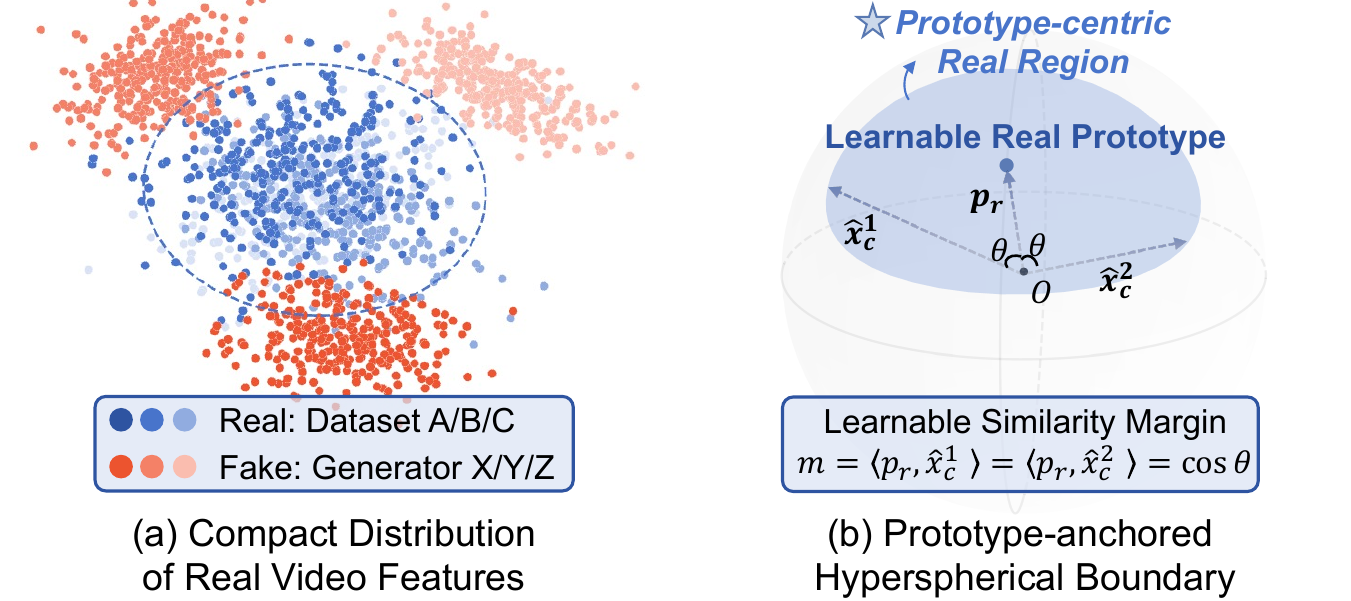} 
  \setlength{\abovecaptionskip}{-3pt} 
  \setlength{\belowcaptionskip}{-5pt}
    \caption{Illustration of our first observation and corresponding design.
    \normalfont
    (a) Real features exhibit a compact distribution across datasets, whereas fake samples form distinct clusters due to generative diversity.
    (b) Real features are constrained within a prototype-centric real region, thereby forming the Prototype-anchored Hyperspherical Boundary for classification.
    }
  \Description{Motivation of our framework.}
  \label{fig:motivation}
\end{figure}
In recent years, video generation models~\cite{ma2025controllable} have advanced rapidly and attracted widespread attention. While these models offer high commercial value, they also raise serious security concerns, such as the spread of misinformation. Therefore, AI-generated video (AIGV) detection, which aims to distinguish real videos from AI-generated (or simply ``fake'') ones, has become increasingly important.

As the quality of AI-generated videos continues to improve, distinguishing them from real videos has become increasingly challenging. Existing AIGV detectors mainly rely on capturing implicit artifacts, i.e., model-specific forgery cues. However, detectors trained on existing data often fail to generalize to newly emerging generative models, as different models introduce distinct artifacts. Compared to repeatedly retraining models, continual learning (CL) provides a more practical solution, enabling models to incrementally adapt to new artifacts while preserving previously learned knowledge~\cite{wang2024comprehensive,cheng2025stacking}. Although CL has shown promising results in AI-generated image detection~\cite{zhang2025devfd,hu2025saido,tian2024dynamic,zhang2025generalization}, its exploration in AIGV detection remains limited.
 
To address this gap, we propose a novel continual learning framework, \textbf{SphereVideo}, which introduces a hyper\underline{sphere}-based decision boundary for AI-generated \underline{video} detection. Specifically, the SphereVideo framework is motivated by two key observations.

First, as shown in Fig.~\ref{fig:motivation}a, \textbf{features of real samples from different datasets tend to cluster within a compact region in the feature space}, whereas fake samples generated by different generators form distinct clusters, despite the lack of any explicit guidance. This is because real videos contain no artifacts, while AI-generated videos exhibit model-specific artifacts due to heterogeneous generation mechanisms~\cite{tang2025towards,li2021frequency}.
This implies that, although future generative artifacts are difficult to anticipate, the stable feature distribution of real videos can serve as a reliable anchor for continual learning.
Building upon this observation and further inspired by Liu et al.~\cite{liu2017sphereface}, feature magnitude is often influenced by task-irrelevant factors such as illumination variations and compression distortions, making it less reliable for discrimination. This motivates us to focus on feature direction as the primary discriminative cue, enabling the model to learn more robust and invariant representations.

Based on these insights, we propose a \textbf{Prototype-anchored Hyperspherical Boundary}, as shown in Fig.~\ref{fig:motivation}b. By normalizing features to discard magnitude and focus on feature direction, we obtain a hypersphere.
Real data is distributed within a region on the hypersphere, and we denote the center of this region as the real prototype. 
Subsequently, we maximize the cosine similarity between the prototype and any real sample to learn a similarity margin. 
This margin shapes a bounded hyperspherical cap, referred to as the prototype-centric real region (or simply the ``real region'').
Meanwhile, features of fake samples are constrained to lie outside this region, with cosine similarity below the margin.
Therefore, an explicit and controllable decision boundary is established on the hypersphere.
With this design, our framework establishes a stable geometric anchor for CL by pulling real samples toward the learnable prototype while repelling fake samples.
As a result, the model consistently preserves the intrinsic characteristics of real data and effectively separates AI-generated patterns.
In contrast to traditional CL methods that often suffer from severe decision boundary shifts when encountering new data, our method explicitly regulates the boundary based on the stable real data distribution, thereby effectively mitigating catastrophic forgetting.

\begin{figure}[t]  
  \centering
  \includegraphics[width=\linewidth]{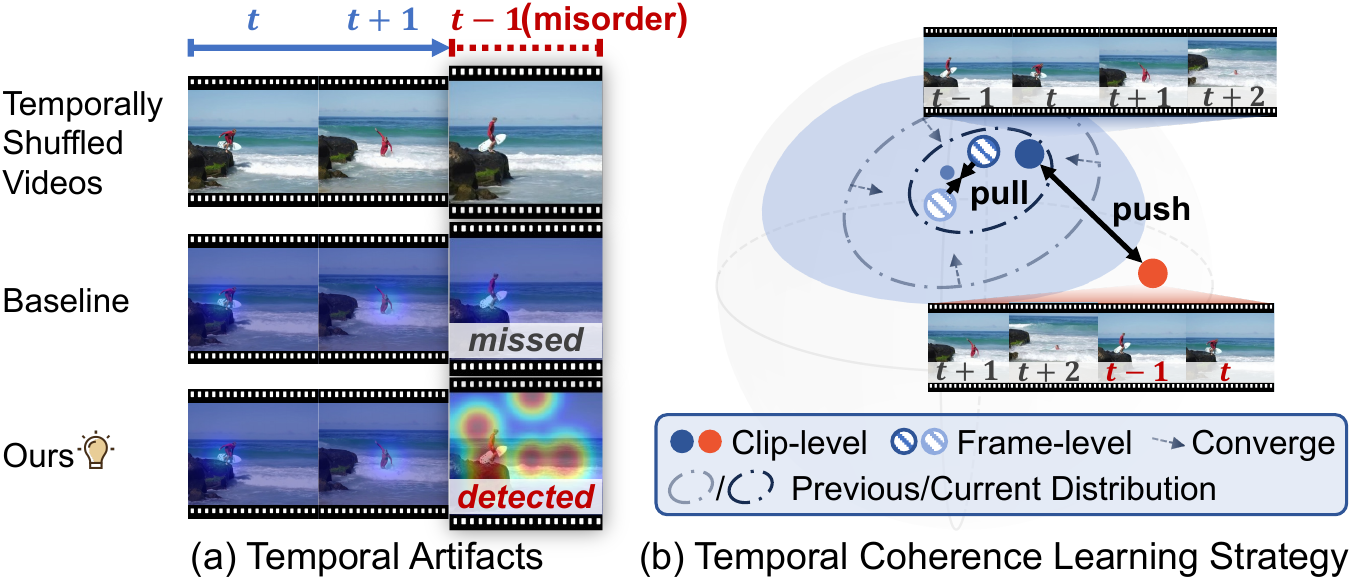} 
  \setlength{\abovecaptionskip}{-3pt}
  \setlength{\belowcaptionskip}{-5pt}
  \caption{Illustration of our second observation and corresponding design.
\normalfont (a) Baselines often rely on spatial shortcuts, failing to capture temporal incoherence.
(b) For real clips, adjacent frame features are pulled together, while original and shuffled clip features are pushed apart, encouraging more compact distributions.}
  \Description{Illustration of our design.}
  \label{fig:contribution}
\end{figure}

Second, we observe that \textbf{existing methods tend to rely on spatial artifacts as shortcuts}. In AIGV, artifacts can be categorized into spatial artifacts (intra-frame inconsistencies) and temporal artifacts (inter-frame incoherence). However, existing methods tend to overfit spatial artifacts while neglecting temporal information~\cite{zheng2021exploring}. 
As shown in Fig.~\ref{fig:contribution}a, when processing a temporally misordered frame in a real video, the activation maps of these methods exhibit no strong response, indicating their insensitivity to temporal incoherence.
Meanwhile, current generative models produce increasingly photorealistic individual frames, making spatial artifacts harder to perceive. Therefore, joint modeling of spatial and temporal artifacts is important for reliable detection. 

To this end, we propose a \textbf{Temporal Coherence Learning Strategy} to enhance temporal modeling. As shown in Fig.~\ref{fig:contribution}b, we introduce two complementary constraints targeting real data.
First, a frame-level temporal smoothness constraint encourages consistency of feature directions across adjacent real frames. Second, a clip-level shuffle contrastive constraint distinguishes real clips from temporally shuffled ones, where shuffled clips disrupt global temporal coherence while preserving local structure. This design encourages the model to capture temporal incoherence rather than relying solely on spatial cues, and further strengthens real data modeling. 
As a result, the model learns a more representative real prototype and regularizes real data features to be compactly distributed within the real region, forming a stable geometric boundary for classification. Compared with existing methods for temporal modeling, including spatial kernel constraints~\cite{zheng2021exploring} and alternating training~\cite{wang2023altfreezing}, our strategy is better aligned with the boundary design, forming a unified framework.

Along with the SphereVideo framework, we construct a challenging benchmark for continual AIGV detection, with two distinct experimental protocols enabling comprehensive evaluation from multiple perspectives. Extensive results further demonstrate the superiority of SphereVideo.
In summary, our key contributions are:
\begin{itemize}[nosep, leftmargin=*, topsep=0pt, partopsep=0pt]
\item We propose a Prototype-anchored Hyperspherical Boundary that defines a compact region representing the distribution of real data features. The region is centered around a real prototype, which serves as a stable anchor for continual learning to mitigate representation drift and catastrophic forgetting.
\item To mitigate shortcut learning on spatial artifacts, we impose temporal learning constraints on real data at different feature levels, enhancing temporal modeling while promoting the learning of a real prototype and the formation of a stable decision boundary.
\item We establish a new benchmark, and SphereVideo achieves outstanding performance under two protocols, with an optimal plasticity-stability trade-off and superior generalization.
\end{itemize}

\section{Related Work}
\subsection{AI-generated Video Detection}
Recent studies in AI-generated video detection \cite{chen2024demamba, ni2025genvidbench, zheng2025d3, chen2025genworld, liu2025lavid, zhangphysics, corviseeing, internoai, lipreserving} have focused on uncovering artifacts in AIGVs, with the primary goal of enhancing generalization to unseen data. 
For example, Vahdati et al. \cite{vahdati2024beyond} show that video generators leave distinct temporal artifacts compared to image generators, providing a foundation for AIGV detection. Song et al. \cite{song2024learning} leverage Large Multi-modal Models to capture semantic artifacts and utilize a VQ-VAE reconstruction process to amplify diffusion-specific features. 
Furthermore, He et al. \cite{he2024exposing} propose a dual-stream framework that adaptively integrates local motion information and global appearance variations to expose subtle temporal cues.
In parallel, extensive research has targeted deepfake video detection \cite{zheng2021exploring, wang2023altfreezing, cozzolino2021id, xu2023tall, dong2022protecting,zhang2024learning}, primarily focusing on facial artifacts. For instance, Choi et al. \cite{choi2024exploiting} use contrastive learning to model unnatural temporal variations in style latent vectors. Yan et al. \cite{yan2025generalizing} identify a universal temporal artifact called Facial Feature Drift and propose a video-level blending method to simulate it.
Additionally, Ciamarra et al. \cite{ciamarra2024temporal} use temporal surface frames to model pixel-level dynamics, identifying deepfakes through temporal jitters. 
Despite their effectiveness, these methods rely on limited data, making it impractical to train a universally generalizable detector under rapidly evolving generative models. Continual learning, which enables adaptation to new data without catastrophic forgetting, provides a more practical solution.

\subsection{AI-generated Image Detection under Continual Learning}
Compared to AIGV detection, AI-generated image detection focuses solely on spatial artifacts. Recent studies have explored continual learning in this setting to address the plasticity-stability trade-off.
Some works emphasize modularity to isolate task-specific features. For instance, Zhang et al.~\cite{zhang2025devfd} introduce a developmental Mixture-of-Experts (MoE) framework that decouples real and fake face modeling via LoRA modules, while Hu et al.~\cite{hu2025saido} leverage multimodal large language models for scene-aware expert allocation, further enhanced by importance-aware gradient projection.
Some works focus on experience replay and prototype learning to reconstruct historical distributions. Tian et al.~\cite{tian2024dynamic} propose Dynamic Mixed Prototypes (DMP) for prototype-based replay and representation distillation, while Cheng et al.~\cite{cheng2025stacking} employ Sparse Uniform Replay (SUR) to isolate and align feature distributions across diverse forgery types.
Other works impose constraints on the feature space and optimization process.
Zhang et al.~\cite{zhang2025generalization} utilize learnable watermarks in hyperbolic space for task alignment, Yang et al.~\cite{yanghsic} construct an information bottleneck via the Hilbert-Schmidt Independence Criterion (HSIC) to suppress semantic noise, Tang et al.~\cite{tang2025towards} reduce domain shift through content-agnostic adapters, and Wang et al.~\cite{wang2026generalizable} integrate progressive augmentation, K-FAC optimization, and linear mode connectivity to enhance robust generalization via invariant feature alignment.
Despite progress in image-level tasks, continual AIGV detection remains limited exploration. Compared with images, videos introduce additional temporal artifacts, making detection more challenging. To address this gap, we propose a novel framework and further introduce the first comprehensive benchmark for continual AIGV detection.

\section{Proposed SphereVideo Method}
\begin{figure*}[t] 
  \centering
  \includegraphics[width=1\textwidth]{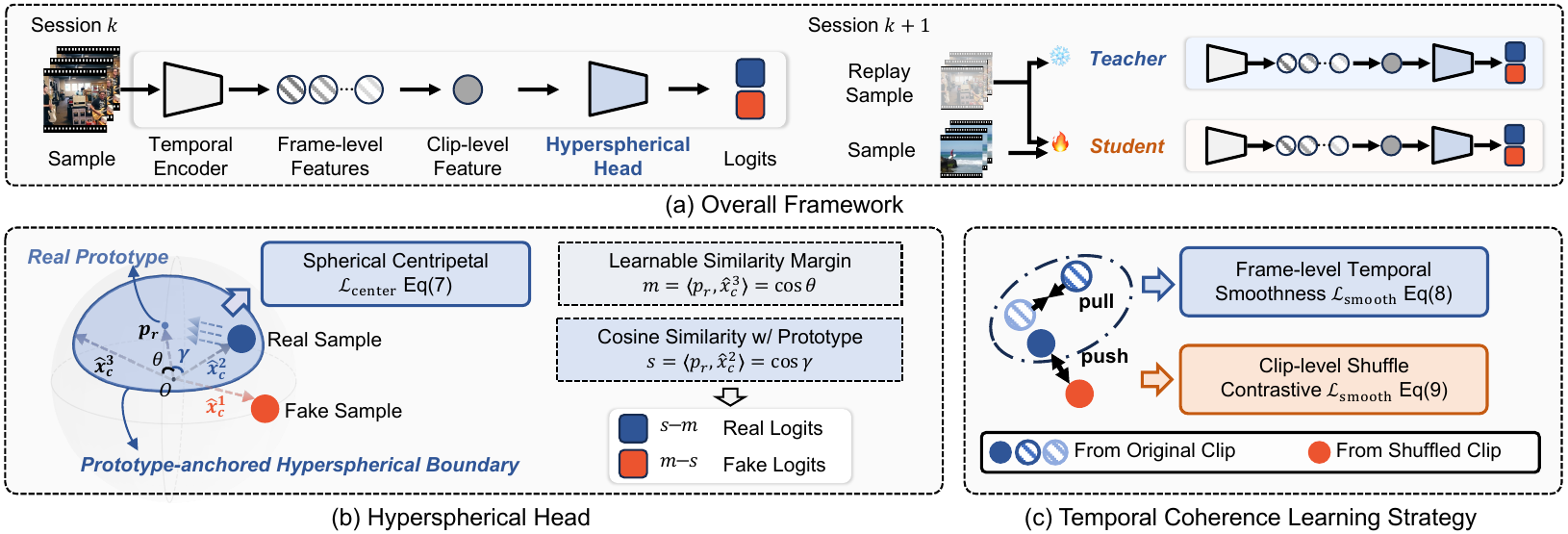} 
  \setlength{\abovecaptionskip}{-5pt}
  \setlength{\belowcaptionskip}{-5pt}
  \caption{Overview of our SphereVideo framework. \normalfont
  (a) Input samples are processed through a temporal encoder to extract frame-level features, which are aggregated into clip-level feature to derive logits via hyperspherical head, followed by softmax-based prediction. Replay sample is used to preserve prior knowledge in new session. (b) The hyperspherical head derives logits via a Prototype-anchored Hyperspherical Boundary and employs a centripetal loss to facilitate real feature compactness. (c) The Temporal Coherence Learning Strategy pulls together features of adjacent real frames while pushing apart those from original and shuffled clips.}
  \label{fig:framework}
\end{figure*}
\subsection{Preliminary of Continual Learning}
\textbf{Training Paradigm.} In continual learning, new data is introduced sequentially to fine-tune a model that has already been trained on prior sessions, and the complete prior data remains inaccessible\cite{de2021continual}.

\noindent\textbf{Research Objective.} A key challenge in continual learning is catastrophic forgetting~\cite{sun2025continual}, where performance on previous sessions degrades as the model adapts to new ones. Accordingly, our goal is to develop a framework that accurately classifies videos from both past and current sessions as real or fake.

\noindent\textbf{Replay and Distillation Methods.} Experience replay (ER) is a widely used strategy for mitigating catastrophic forgetting~\cite{wang2024comprehensive}. After each session, a subset of past samples is stored in a memory buffer and jointly used with new data to update the model. Moreover, ER can be naturally combined with knowledge distillation, where a frozen model from previous sessions acts as a teacher to guide the student via replayed samples, thereby facilitating knowledge transfer. Our method employs these strategies.

\subsection{Overview}
Formally, the training dataset at the $k$-th session is defined as 
$D = \{(c^i, y^i)\}_{i=1}^{N + kM}$, where $k \geq 0$ denotes the session index. 
Here, $c^i \in \mathbb{R}^{T \times H \times W}$ denotes the i-th clip, and $y^i \in \{0,1\}$ represents its label, where $0$ indicates real and $1$ indicates fake. 
$N$ denotes the number of newly introduced samples in the current session, and $M$ denotes the number of replay samples from each previous session.
Each clip is sampled from a video and assigned the corresponding video label.
During inference, video-level predictions are obtained by averaging the prediction scores of all sampled clips. 
In the following, we treat clips as the basic units for model description.

As illustrated in Fig.~\ref{fig:framework}a, given an input clip $c^i$, we employ a temporal encoder $\mathcal{E}(\cdot)$ to extract frame-level features 
$\mathbf{x}_f \in \mathbb{R}^{\frac{T}{2} \times d}$, where $\frac{T}{2}$ is determined by the encoder (detailed in Sec.~\ref{subsec:temporal_encoder}) and $d$ denotes the feature dimension. 
The frame-level features are then aggregated along the temporal dimension to obtain clip-level feature $\mathbf{x}_c \in \mathbb{R}^d$, which is subsequently processed by the Hyperspherical Head (Fig.~\ref{fig:framework}b) for classification. 
Specifically, the head  $\ell_2$-normalizes $\mathbf{x}_c$ onto a hypersphere to obtain $\hat{\mathbf{x}}_c$.
On the hypersphere, we define a compact real region centered around a learnable real prototype $\mathbf{p}_r$  to model the real feature distribution, which further establishes a Prototype-anchored Hyperspherical Boundary with a similarity margin $m$.
The logits for real and fake classes are computed based on the cosine similarity $s$ between $\hat{\mathbf{x}}_c$ and $\mathbf{p}_r$ and the margin $m$. In particular, if $s - m > 0$, the feature lies within the real region, yielding a higher logit for the real class. Finally, a softmax function produces a prediction score $p \in [0, 1]$, representing the probability of the clip being AI-generated.

Furthermore, to enhance temporal modeling and improve the compactness of real features, we introduce a Temporal Coherence Learning Strategy (Fig.\ref{fig:framework}c). It regularizes temporal variations between adjacent real frames to be smooth, while pushing real clip features away from their shuffled counterparts in the feature space.

\subsection{Temporal Encoder}
\label{subsec:temporal_encoder}
We adopt the pretrained VideoMAE~\cite{tong2022videomae} as our temporal encoder $\mathcal{E}(\cdot)$, as it learns spatiotemporal representations from video data via self-supervised reconstruction. This enables VideoMAE to effectively capture the intrinsic structure of real videos, thereby facilitating the compactness of the real region in our framework.

Specifically, VideoMAE employs joint spatiotemporal patch embedding, treating each $2 \times 16 \times 16$ patch as a visual token.
For an input video clip of size $T \times H \times W$, the model generates a sequence of $\frac{T}{2} \times \frac{H}{16} \times \frac{W}{16}$ visual tokens, each represented as a $d$-dimensional vector.
The resulting token-level features $\mathbf{z}$ are:
\begin{equation}
\mathbf{z} = \mathcal{E}(c^i) \in \mathbb{R}^{\frac{T}{2} \times \frac{H}{16} \times \frac{W}{16} \times d}.
\label{eq:token-level}
\end{equation}

At each temporal step $t$, frame-level features $\mathbf{x}_f$ are the average of spatial tokens $\mathbf{z}[t, p]$, where $p \in [1, \frac{H}{16}\cdot\frac{W}{16}]$:
\begin{equation}
\mathbf{x}_f[t] = \frac{1}{\frac{H}{16} \cdot \frac{W}{16}} \sum_{p=1}^{\frac{H}{16}\cdot\frac{W}{16}} \mathbf{z}[t, p].
\end{equation}

Furthermore, the clip-level feature is the temporal average of $\mathbf{x}_f$, used for classification:
\begin{equation}
\mathbf{x}_c = \frac{2}{T} \sum_{t=1}^{T/2 } \mathbf{x}_f[t].
\end{equation}

\subsection{Prototype-anchored Hyperspherical Boundary}
\label{sec:Boundary}
Features are normalized to a hypersphere where we learn a compact real region centered at a real prototype. This defines a decision boundary: Clips are classified as real if their deviation from the prototype lies within this region, and AI-generated otherwise.

Specifically, we employ a hyperspherical head $\mathcal{H}(\cdot)$ (Fig.~\ref{fig:framework}b) that $\ell_2$-normalizes clip features $\mathbf{x}_c$ onto a hypersphere to obtain $\hat{\mathbf{x}}_c$, effectively discarding magnitude variations to focus on directional information.
A learnable real prototype $\mathbf{p}_{r} \in \mathbb{R}^d$ (where $\|\mathbf{p}_{r}\|_2 = 1$) is optimized on the hypersphere as the central anchor.
This head $\mathcal{H}(\cdot)$ distinguishes between real and fake clips by measuring the similarity between $\hat{\mathbf{x}}_c$ and the real prototype $\mathbf{p}_{r}$. Formally, the cosine similarity $s \in [-1, 1]$ is computed as follows:
\begin{equation}
    s = \langle \hat{\mathbf{x}}_c, \mathbf{p}_{r} \rangle, \quad \hat{\mathbf{x}}_c = \mathbf{x}_c / \|\mathbf{x}_c\|_2,
\end{equation}
where $\|\cdot\|_2$ and $\langle \cdot, \cdot \rangle$ denote the $\ell_2$ norm and inner product, respectively. 
To define a decision boundary, we introduce a learnable similarity margin $m \in [0, 1)$ as a threshold on cosine similarity. 
The relative difference between $s$ and $m$ determines the class logits. Specifically, the logits $\mathbf{l} = [l_0, l_1]$, corresponding to the real and fake classes, are computed as:
\begin{equation}
\mathbf{l} = \big[\, s - m, \, m - s \,\big] \cdot \alpha,
\label{eq:logits}
\end{equation}
where $\alpha$ is a scaling factor to stabilize training. Consequently, features closer to $\mathbf{p}_{r}$ yield higher real-class logits, while those farther away yield higher fake-class logits.
Finally, these logits are converted to class probabilities via the softmax function:
\begin{equation}
p = \text{softmax}(\mathbf{l})[1] 
= \frac{\exp(l_1)}{\exp(l_0) + \exp(l_1)},
\label{eq:pre}
\end{equation}
where $p \in [0, 1]$ represents the predicted probability of the clip being AI-generated. 
Ideally, $p$ approaches $0$ for real clips and $1$ for AI-generated ones. We adopt a classification threshold of $0.5$, identifying clips as AI-generated if $p > 0.5$, and real otherwise.

Geometrically, the learnable similarity margin $m$ defines a spherical cap centered at $\mathbf{p}_{r}$ on the hypersphere, 
effectively forming a clear decision boundary. Features closer to $\mathbf{p}_{r}$ that fall within this spherical cap are identified as real, while those falling outside this region are classified as AI-generated.

We further define a spherical centripetal loss $\mathcal{L}_{\text{center}}$ to encourage real clips to cluster around the real prototype:
\begin{equation}
\mathcal{L}_{\text{center}} = \frac 1 {N_{\text{real}}} \sum_{i=1}^{N_{\text{real}}} \left( 1 - \langle \hat{\mathbf{x}}_c^i, \mathbf{p}_{r} \rangle \right)^2,
\label{eq:center_loss}
\end{equation}
where $N_{\text{real}}$ denotes the number of real clips, and $\hat{\mathbf{x}}_c^i$ is the $\ell_2$-normalized feature of the $i$-th real video clip. The loss penalizes large deviations, encouraging real clip features to concentrate around the prototype and form a compact real region.

\subsection{Temporal Coherence Learning Strategy} 
\label{sec:temporal_learning}
To improve the model’s capacity for capturing temporal dynamics, we propose a Temporal Coherence Learning Strategy (Fig.\ref{fig:framework}c), which imposes two complementary constraints.

\textbf{Frame-level Temporal Smoothness Loss.} 
For real video clips, the evolution of features across temporally adjacent frames is expected to be smooth. 
Let $\hat{\mathbf{x}}_f^{i,t} \in \mathbb{R}^d$ denote the $\ell_2$-normalized feature of the $t$-th frame in the $i$-th real clip. 
The frame-level temporal smoothness loss $\mathcal{L}_{\text{smooth}}$ is defined as:
\begin{equation}
\mathcal{L}_{\text{smooth}} = \frac{1}{N_{real} (T/2-1)} \sum_{i=1}^{N_{real}} \sum_{t=1}^{T/2-1} \Big( 1 - \langle \hat{\mathbf{x}}_f^{i,t},  \hat{\mathbf{x}}_f^{i,t+1} \rangle \Big ).
\label{eq:smooth_loss}
\end{equation}

\textbf{Clip-level Shuffle Contrastive Loss.} 
To explicitly model temporal order at the clip level, we randomly partition each real clip into $L$ consecutive temporal segments and then shuffle these segments to generate negative examples. This segment-level shuffling preserves local temporal structure while disrupting global consistency, thereby forcing the model to capture finer-grained temporal patterns.
Let $\hat{\mathbf{x}}_c^i \in \mathbb{R}^d$ denote the normalized clip-level feature of the $i$-th real clip, and $\hat{\tilde{\mathbf{x}}}_c^i$ denote the corresponding feature after segment shuffling. 
The contrastive loss $\mathcal{L}_{\text{shuffle}}$ is formulated as:
\begin{equation}
\mathcal{L}_{\text{shuffle}} = \frac{1}{N_{real}} \sum_{i=1}^{N_{real}} \Big( 1 + \langle \hat{\mathbf{x}}_c^i , \hat{\tilde{\mathbf{x}}}_c^i \rangle \Big).
\label{eq:shuffle_loss}
\end{equation}

The overall temporal coherence loss $\mathcal{L}_{\text{temporal}}$ is then defined as:
\begin{equation}
\mathcal{L}_{\text{temporal}} = \mathcal{L}_{\text{smooth}} + \mathcal{L}_{\text{shuffle}}.
\end{equation}
These two complementary losses penalize abrupt changes between adjacent frames and high similarity between original and shuffled clips respectively. Operating at different levels, they encourage the model to capture the intrinsic dynamics of real clips, thereby promoting more compact real region.

\subsection{Fundamental Loss Functions}
Beyond the aforementioned constraints, several loss functions are essential for continual AI-generated video detection.

\textbf{Cross-Entropy Loss.} 
For classification, we use the standard binary cross-entropy loss $\mathcal{L}_{\text{ce}}$:
\begin{equation}
\mathcal{L}_{\text{ce}} = - \frac{1}{N_{\text{total}}} \sum_{i=1}^{N_{\text{total}}} \Big[ y^i \log p^i + (1 - y^i) \log (1 - p^i) \Big],
\end{equation}
where $p^i$ denotes the predicted probability of the $i$-th clip in Eq.~\eqref{eq:pre}, $y^i \in \{0,1\}$ is its ground truth label, and $N_{\text{total}}$ denotes the total number of samples in the batch, including those from the current session and the replay buffer.

\textbf{Knowledge Distillation Loss.} 
Following Pan et al.~\cite{pan2023dfil}, we employ feature distillation $\mathcal{L}_{\text{fd}}$ and soft-label distillation $\mathcal{L}_{\text{kd}}$ to align the current student model with the frozen teacher model from previous session, thereby mitigating catastrophic forgetting
Specifically, the distillation losses are defined as follows:
\begin{equation}
\mathcal{L}_{\text{fd}} = \frac{1}{N_{\text{replay}}} \sum_{i=1}^{N_{\text{replay}}} \left\| \hat{\mathbf{z}}_{\text{stu}}^{i} - \hat{\mathbf{z}}_{\text{tea}}^{i} \right\|_2^2,
\end{equation}
\begin{equation}
\mathcal{L}_{\text{kd}} = \frac{\tau^2}{N_{\text{replay}}} \sum_{i=1}^{N_{\text{replay}}} 
\mathrm{KL}\Big( \text{softmax}(\mathbf{I}_{\text{tea}}^{i} / \tau) \, \| \, \text{softmax}(\mathbf{I}_{\text{stu}}^{i} / \tau) \Big),
\label{eq:kdloss}
\end{equation}
where $N_{\text{replay}}$ is the number of replay samples, $\tau$ is the temperature, and $\mathrm{KL}(\cdot \| \cdot)$ denotes the Kullback--Leibler divergence. $\hat{\mathbf{z}}_{\text{stu}}^{i}, \hat{\mathbf{z}}_{\text{tea}}^{i}$ and $\mathbf{I}_{\text{stu}}^{i}, \mathbf{I}_{\text{tea}}^{i}$denote the $\ell_2$-normalized token features and logits for the student and teacher models, respectively, as defined in Eqs.~\eqref{eq:token-level} and ~\eqref{eq:logits}.
Then, the overall distillation loss $\mathcal{L}_{\text{dist}}$ is defined as:
\begin{equation}
\mathcal{L}_{\text{dist}} = \mathcal{L}_{\text{fd}} + \mathcal{L}_{\text{kd}}.
\end{equation}

\textbf{Overall Loss.} 
The final loss function for continual learning integrates all components as follows:
\begin{equation}
\mathcal{L}_{\text{total}} = \mathcal{L}_{\text{ce}} + \mathcal{L}_{\text{dist}} + \mathcal{L}_{\text{temporal}} + \mathcal{L}_{\text{center}},
\end{equation}
where $\mathcal{L}_{\text{ce}}$ denotes the primary classification loss, and $\mathcal{L}_{\text{dist}}$ mitigates catastrophic forgetting via distillation. $\mathcal{L}_{\text{temporal}}$ enhances temporal modeling, and together with $\mathcal{L}_{\text{center}}$, encourages more compact feature distributions, thereby facilitating a stable decision boundary. Assigned equal weights, these complementary components jointly enable effective continual learning.

\section{Experiments}
\begin{table*}[t]
\centering
\small
\setlength{\aboverulesep}{0pt}
\setlength{\belowrulesep}{0pt}
\setlength{\tabcolsep}{2pt}
\renewcommand{\arraystretch}{1.2}
\setlength{\abovecaptionskip}{7pt}
\caption{Performance comparison with state-of-the-art methods across nine sessions under protocol \ref{prot:Protocol1} (\%). 
\normalfont
Numerical prefixes (e.g., "0-") denote the session index. Bold and underlined values represent the best and second-best results, respectively. Our SphereVideo (highlighted in blue) achieves a superior stability-plasticity trade-off.}
\label{tab:CL-result}

\newcolumntype{Y}{>{\centering\arraybackslash}X}
\newcolumntype{H}{>{\hsize=1.9\hsize\centering\arraybackslash}X}
\newcolumntype{S}{>{\hsize=0.85\hsize\centering\arraybackslash}X}
\begin{tabularx}{\dimexpr\textwidth-5pt\relax}{ l ccc c H c SS c SS c SS }
\toprule
\multirow{2}{*}{\textbf{Method}} 
& \multirow{2}{*}{\makecell{\textbf{Conference}\\\textbf{\& Year}}}
& \multirow{2}{*}{\makecell{\textbf{Continual}\\\textbf{Learning}}}
& \multirow{2}{*}{\makecell{\textbf{Replay}\\\textbf{Set}}}
& & \textbf{0-HunyuanVideo}
& & \multicolumn{2}{c}{\textbf{1-EasyAnimate}}
& & \multicolumn{2}{c}{\textbf{2-CogVideoX}}
& & \multicolumn{2}{c}{\textbf{3-LTX-Video}} \\
\cmidrule(lr){6-6} \cmidrule(lr){8-9} \cmidrule(lr){11-12} \cmidrule(lr){14-15}
& & & & & AA & & AA & AF & & AA & AF & & AA & AF \\
\midrule
DeMamba~\cite{chen2024demamba} & arXiv 2024 & $\times$ & $\times$ & & 98.50 & & 97.50 & 3.00 & & 97.17 & 2.50 & & 94.87 & 4.33 \\
ReStraV~\cite{interno2025ai} & NeurIPS 2025 & $\times$ & $\times$ & & 83.50 & & 50.25 & 33.50 & & 50.23 & 32.00 & & 50.88 & 21.83 \\
\midrule
DFIL~\cite{pan2023dfil} & ACMMM 2023 & $\checkmark$ & $\checkmark$ & & 97.50 & & 96.00 & \underline{0.00} & & 94.33 & \textbf{$-$1.25} & & 96.75 & \textbf{$-$2.67} \\
SUR-LID~\cite{cheng2025stacking} & CVPR 2025 & $\checkmark$ & $\checkmark$ & & \underline{98.00} & & \textbf{99.00} & \underline{0.00} & & \textbf{98.33} & 0.50 & & \underline{97.00} & 1.17 \\
Wang et al.~\cite{wang2026generalizable} & TMM 2025 & $\checkmark$ & $\checkmark$ & & 87.00 & & 87.25 & 1.50 & & 86.50 & \underline{0.00} & & 86.63 & \underline{$-$0.05} \\
Tang et al.~\cite{tang2025towards} & TIFS 2025 & $\checkmark$ & $\checkmark$ & & \textbf{98.50} & & \underline{98.25} & 1.00 & & \underline{97.67} & 0.25 & & 96.12 & 1.50 \\
\rowcolor{softblue}
\textbf{SphereVideo (ours)} & - & $\checkmark$ & $\checkmark$ & & 97.50 & & \textbf{99.00} & \textbf{$-$0.50} & & \textbf{98.33} & 0.25 & & \textbf{97.38} & 0.50 \\
\bottomrule
\end{tabularx}

\vspace{1.5mm}

\begin{tabularx}{\dimexpr\textwidth-5pt\relax}{ l c YY c YY c YY c YY c YY c Y }
\toprule
\multirow{2}{*}{\textbf{Method}} 
& & \multicolumn{2}{c}{\textbf{4-Magi-1}}
& & \multicolumn{2}{c}{\textbf{5-CausVid}}
& & \multicolumn{2}{c}{\textbf{6-ZeroScope}}
& & \multicolumn{2}{c}{\textbf{7-VideoCrafter1}}
& & \multicolumn{2}{c}{\textbf{8-SVD}}
& & \multirow{2}{*}{\makecell{\textbf{New.}\\\textbf{Acc}}} \\
\cmidrule(lr){3-4} \cmidrule(lr){6-7} \cmidrule(lr){9-10} \cmidrule(lr){12-13} \cmidrule(lr){15-16}
& & AA & AF & & AA & AF & & AA & AF & & AA & AF & & AA & AF & & \\
\midrule
DeMamba~\cite{chen2024demamba}  & & 84.00 & 17.37 & & 77.92 & 23.20 & & 58.16 & 45.33 & & 71.78 & 28.00 & & 64.02 & 36.57 & & 96.53 \\
ReStraV~\cite{interno2025ai}  & & 65.30 & 2.87 & & 50.00 & 18.20 & & 50.00 & 15.33 & & 50.00 & 13.41 & & 58.00 & 4.92 & & 55.44 \\
\midrule
DFIL~\cite{pan2023dfil}  & & 93.30 & \underline{1.37} & & 88.50 & \underline{2.70} & & 89.27 & \underline{0.75} & & \underline{88.52} & \underline{2.89} & & 86.70 & 6.91 & & 87.65 \\
SUR-LID~\cite{cheng2025stacking}  & & 92.20 & 6.75 & & 88.00 & 11.30 & & \underline{90.41} & 8.08 & & 88.31 & 9.43 & & 86.87 & 11.27 & & \textbf{96.89} \\
Wang et al.~\cite{wang2026generalizable} & & 82.80 & \textbf{0.25} & & 78.58 & \textbf{$-$1.60} & & 76.52 & \textbf{$-$1.00} & & 75.80 & \textbf{$-$0.14} & & 76.32 & \textbf{$-$1.62} & & 72.43 \\
Tang et al.~\cite{tang2025towards} & & \underline{94.00} & 2.37 & & \underline{91.25} & 4.50 & & 88.96 & 6.75 & & 87.90 & 6.36 & & \underline{87.99} & 6.41 & & 93.08 \\
\rowcolor{softblue}
\textbf{SphereVideo (ours)} & & \textbf{94.40} & 3.75 & & \textbf{91.67} & 6.30 & & \textbf{91.93} & 5.92 & & \textbf{90.84} & 6.36 & & \textbf{91.07} & \underline{6.32} & & \underline{96.64} \\
\bottomrule
\end{tabularx}
\end{table*}

\subsection{Experimental Settings}
\label{Sec:ExperimentalSetting}
\textbf{Datasets.}
We construct a diverse benchmark covering both classical and state-of-the-art video generation methods.
Existing methods can be broadly categorized into three groups: UNet-based diffusion, DiT-based diffusion, and autoregressive models~\cite{ma2025controllable}. Accordingly, our benchmark includes all three categories for comprehensive and challenging evaluation.
Specifically, UNet-based diffusion models include SVD~\cite{blattmann2023stable}, VideoCrafter1~\cite{chen2023videocrafter1}, Zeroscope from DVF~\cite{song2024learning}, and GenVideo~\cite{chen2024demamba}. DiT-based diffusion models include EasyAnimateV5.1~\cite{xu2024easyanimate}, HunyuanVideo~\cite{kong2024hunyuanvideo}, CogVideoX~\cite{yang2024cogvideox}, and LTX-Video~\cite{hacohen2024ltx} from GenBuster-200K~\cite{wen2025busterx}. We further include autoregressive models such as Magi-1~\cite{teng2025magi} and CausVid~\cite{yin2025slow}, for which we generate videos using text-to-video pipelines with prompts from real datasets.
For real videos, we integrate HD-VG-130M~\cite{wang2023videofactory}, YouTube-8M~\cite{abu2016youtube}, InternVid-10M~\cite{wang2023internvid}, and OpenVid-1M~\cite{nan2024openvid}, covering diverse content, resolutions, frame rates, and durations to reduce dataset bias. 
Furthermore, we adopt the open-world benchmark from GenBuster-200K~\cite{wen2025busterx}, which includes eight commercial AIGV models (e.g., Sora~\cite{brooks2024video} and Pika) to evaluate generalization.

\noindent\textbf{Evaluation Metrics.}
We report Accuracy (Acc), Average Accuracy (AA), Average Forgetting (AF), New Accuracy (New.Acc), and mean Average Accuracy (mAA). Acc denotes the accuracy on the corresponding test sets after the final session. AA and AF assess overall performance and memory stability during continual learning, where AA averages the accuracies across all learned sessions and AF quantifies the performance degradation on past sessions. New.Acc denotes the average accuracy on newly introduced data across all sessions, reflecting the model’s plasticity. mAA denotes the mean of AA values computed at each session.

\noindent\textbf{Protocols.}
To evaluate performance in continual learning settings and generalization to unseen data in open-world scenarios, we design two distinct protocols. 

\begin{itemize}[leftmargin=*, topsep=0pt, itemsep=0pt, parsep=0pt]
    \item \refstepcounter{protocol}\label{prot:Protocol1} \textbf{Protocol \theprotocol.} We construct a sequential dataset 
    \{HunyuanVideo, EasyAnimate, CogVideoX, LTX-Video, Magi-1, CausVid, Zeroscope, VideoCrafter1, SVD\}
    for evaluation in CL settings.

    \item \refstepcounter{protocol}\label{prot:Protocol2} \textbf{Protocol \theprotocol.} We use the open-world benchmark 
    \{Gen-3, Jimeng, Kling, Luma, Pika, Sora, Vidu, Wanx\} from GenBuster-200K~\cite{wen2025busterx}
    to evaluate generalization to unseen data.
\end{itemize}

\noindent\textbf{Implementation Details.}
In our continual learning dataset, Zeroscope and VideoCrafter1 contain 450 and 800 videos, respectively, while all other generative methods include 1,000 videos each. The number of corresponding real videos is aligned with that of the AI-generated videos in each session.
Videos are sampled into frames at their original frame rates to preserve temporal fidelity. Continuous temporal windows are used to extract frames (16 frames for training and 32 for testing). The window is further partitioned into units of length 4. Within each unit, random jitter is applied to the starting index to obtain clips of length $T = 4$, thereby enhancing temporal robustness.
We adopt VideoMAE~\cite{tong2022videomae} pretrained on Kinetics-400 as the backbone ($d=768$). Adam is used with a learning rate of $1\times10^{-5}$ for 5 epochs and a batch size of 32. 
Each session maintains a class-wise replay buffer (64 samples per class), from which 4 samples per class are replayed in each batch. Clips are randomly sampled using a uniform sampling strategy.
We set $\alpha = \tau = 20$ in Eqs.~\eqref{eq:logits} and~\eqref{eq:kdloss}, and $L=2$ in Sec.~\ref{sec:temporal_learning}. 
Training is conducted on four NVIDIA GeForce RTX 4090 GPUs.

\noindent\textbf{Baselines.}
We evaluate our method against state-of-the-art baselines, including conventional AI-generated video detectors~\cite{interno2025ai,chen2024demamba} and image-level CL methods~\cite{pan2023dfil, cheng2025stacking, wang2026generalizable, tang2025towards}. 
For a fair comparison, all baselines are reproduced using the same backbone and experimental configuration as our method.

\begin{table*}[t]
\centering
\small
\setlength{\aboverulesep}{0pt}
\setlength{\belowrulesep}{0pt}
\setlength{\tabcolsep}{2.5pt}
\renewcommand{\arraystretch}{1.2}
\setlength{\abovecaptionskip}{7pt}
\caption{Comparison with state-of-the-art methods under protocol ~\ref{prot:Protocol2} (\%). 
\normalfont
Following the setting of the benchmark~\cite{wen2025busterx}, we report accuracy across OOD generative models, along with accuracy on paired real samples (Real.Acc), overall accuracy on all OOD fake samples (Fake.Acc), and their mean (Avg.Acc).
Results marked with $^*$ are directly taken from the original benchmark without reproduction.}
\label{tab:generalization_result}

\newcolumntype{Y}{>{\centering\arraybackslash}X}

\begin{tabularx}{\dimexpr\textwidth-5pt\relax}{ l c YYYYYYYY c YYY }
\toprule
\multirow{2}{*}{\textbf{Method}} 
& & \multicolumn{8}{c}{\textbf{Out-of-Distribution (OOD) Generative Models (Acc) }} 
& & \multirow{2}{*}{\makecell{\textbf{Real.}\\\textbf{Acc}}} 
& \multirow{2}{*}{\makecell{\textbf{Fake.}\\\textbf{Acc}}} 
& \multirow{2}{*}{\makecell{\textbf{Avg.}\\\textbf{Acc}}} \\
\cmidrule(lr){3-10}
& & \textbf{Gen3} & \textbf{Jimeng} & \textbf{Kling} & \textbf{Luma} & \textbf{Pika} & \textbf{Sora} & \textbf{Vidu} & \textbf{Wanx} & & & & \\
\midrule
BusterX$^*$ \cite{wen2025busterx} & & 82.00 & 82.00 & 81.00 & 82.00 & 81.00 & 81.50 & 82.00 & 88.70 & & $-$ & $-$ & 84.80 \\
\midrule
DeMamba~\cite{chen2024demamba} & & 87.00 & 51.00 & 68.00 & 79.00 & 88.00 & 71.50 & 85.33 & 76.67 & & 66.50 & 75.90 & 71.20 \\
ReStraV~\cite{interno2025ai} & & 37.00 & 63.00 & 73.00 & 49.00 & 84.00 & 50.50 & 74.67 & 78.67 & & 55.00 & 63.70 & 59.35 \\
\midrule
DFIL~\cite{pan2023dfil} & & 87.00 & 52.00 & 81.00 & \underline{87.00} & 93.00 & \underline{84.00} & 94.00 & 91.33 & & \underline{91.60} & 84.60 & \underline{88.10} \\
SUR-LID~\cite{cheng2025stacking} & & \underline{88.00} & \underline{64.00} & \underline{86.00} & 86.00 & \underline{95.00} & 81.50 & 95.33 & \underline{92.00} & & 88.90 & \underline{86.30} & 87.60 \\
Wang et al.~\cite{wang2026generalizable} & & 77.00 & \underline{64.00} & 65.00 & 58.00 & 94.00 & 57.00 & 76.00 & 82.00 & & 89.20 & 70.90 & 80.05 \\
Tang et al.~\cite{tang2025towards} & & 79.00 & \underline{64.00} & 82.00 & 82.00 & 92.00 & 80.50 & \underline{96.67} & 89.33 & & 89.00 & 83.90 & 86.45 \\
\rowcolor{softblue}
\textbf{SphereVideo (ours)} & & \textbf{90.00} & \textbf{74.00} & \textbf{87.00} & \textbf{89.00} & \textbf{97.00} & \textbf{89.00} & \textbf{97.33} & \textbf{94.67} & & \textbf{92.00} & \textbf{90.30} & \textbf{91.15} \\
\bottomrule
\end{tabularx}
\end{table*}

\subsection{Results in Continual Learning Sessions}
We evaluate the continual learning performance using the sequential datasets defined in Protocol ~\ref{prot:Protocol1}, with comparative results summarized in Tab.~\ref{tab:CL-result}.
Notably, non-CL baselines suffer from severe catastrophic forgetting and limited plasticity when encountering novel generation artifacts, highlighting the necessity of the CL paradigm. 
Among CL methods, our SphereVideo framework achieves state-of-the-art performance in Average Accuracy (AA).
In the experimental sequence, the first four, middle two, and final three datasets belong to the same generative model groups, resulting in high similarity of artifacts in the initial four sessions.
Consequently, all methods show comparable performance during these early sessions. 
However, upon the introduction of distinct frameworks starting from \textit{Magi-1}, SphereVideo demonstrates a significant performance improvement, underscoring its robust capability to handle diverse and rapidly evolving artifacts.
Beyond achieving the highest overall AA, our method strikes an optimal balance among AA, Average Forgetting (AF), and New Accuracy (New.Acc). For instance, although SUR-LID~\cite{cheng2025stacking} yields a slightly higher New.Acc (+0.25\%), its high AF reflects severe catastrophic forgetting, compromising its overall AA. Conversely, Tang et al.~\cite{tang2025towards} maintain an AF comparable to ours, yet their New.Acc is 3.56\% lower. Similarly, although DFIL\cite{pan2023dfil} and Wang et al.~\cite{wang2026generalizable} exhibit even lower AF, their New.Acc is significantly inferior to that of our method, leading to poor overall AA. These results collectively underscore their insufficient plasticity in adapting to novel data.
Overall, SphereVideo successfully achieves a favorable stability-plasticity trade-off, demonstrating its superior effectiveness in complex continual learning scenarios.

\subsection{Generalization to Open-world Settings}
We evaluate generalization performance on the open-world benchmark defined in Protocol~\ref{prot:Protocol2}, which comprises samples unseen during training. Results are summarized in Tab.~\ref{tab:generalization_result}. All methods are evaluated after the final session of Protocol~\ref{prot:Protocol1}.

Notably, BusterX~\cite{wen2025busterx} (shown in the first row) represents the original benchmark result. Its training data is limited to the first four models listed in Protocol~\ref{prot:Protocol1}, containing over 100k fake samples.
Despite a considerably smaller training set (fewer than 9k fake samples), several CL methods still outperform BusterX.
This implies that the diversity of generative frameworks encountered during training might be more critical for generalization than the sheer volume of training samples. Such findings further highlight the importance of CL, where models potentially develop stronger generalization by sequentially adapting to diverse generative frameworks, thereby better aligning with real-world scenarios.

Among all methods, our SphereVideo framework demonstrates a clear advantage, achieving state-of-the-art performance across all metrics. Specifically, it surpasses the baselines by 3.05\% in overall accuracy (Avg.Acc) and 4.0\% in accuracy on fake data (Fake.Acc). We attribute this success to the model's ability to capture the intrinsic characteristics of real data, effectively regularizing real features into a compact real region while repelling fake samples. 
Consequently, open-world fake samples are accurately identified as deviating from the learned real distribution, while real samples are mapped into the real region.
This mechanism offers superior controllability and robustness compared to the hyperplane-based decision boundary used in baselines. Hyperplanes merely partition the feature space based on seen data without explicitly modeling the intrinsic nature of samples. 
Such a boundary provides limited control over how unseen data are mapped, making it difficult to predict which side of the hyperplane a novel sample will fall on and often leading to misclassification.
In contrast, SphereVideo learns an explicit real region that serves as a reference, thereby effectively constraining where unseen sample features are distributed in the feature space.

\begin{table*}[t]
\centering
\small
\setlength{\aboverulesep}{0pt}
\setlength{\belowrulesep}{0pt}
\setlength{\extrarowheight}{1pt} 
\renewcommand{\arraystretch}{1.2}
\setlength{\tabcolsep}{3.2pt} 
\setlength{\abovecaptionskip}{7pt}
\caption{Ablation study of the proposed components across nine sessions under protocol \ref{prot:Protocol1} (\%). 
\normalfont
$\mathcal{H}(\cdot)$ denotes the hyperspherical head (Sec.~\ref{sec:Boundary}). $\mathcal{L}_{\text{center}}$, $\mathcal{L}_{\text{smooth}}$, and $\mathcal{L}_{\text{shuffle}}$ are defined in Eqs.~\eqref{eq:center_loss}, \eqref{eq:smooth_loss}, and \eqref{eq:shuffle_loss}. Abbreviations: PHB (Prototype-anchored Hyperspherical Boundary), TCLS (Temporal Coherence Learning Strategy), and shortened dataset names (e.g., Hunyuan, EasyAnim., LTX-Vid., VidCraft1).
}
\label{tab:ablation}

\resizebox{\textwidth}{!}{%
\begin{tabular}{cccc c cc cc cc cc cc cc cc cc}
\toprule
\multicolumn{2}{c}{\textbf{PHB}} & \multicolumn{2}{c}{\textbf{TCLS}} & \textbf{Hunyuan} & \multicolumn{2}{c}{\textbf{EasyAnim.}} & \multicolumn{2}{c}{\textbf{CogVideoX}} & \multicolumn{2}{c}{\textbf{LTX-Vid.}} & \multicolumn{2}{c}{\textbf{Magi-1}} & \multicolumn{2}{c}{\textbf{CausVid}} & \multicolumn{2}{c}{\textbf{ZeroScope}} & \multicolumn{2}{c}{\textbf{VidCraft1}} & \multicolumn{2}{c}{\textbf{SVD}} \\
\cmidrule(lr){1-2} \cmidrule(lr){3-4} \cmidrule(lr){5-5} \cmidrule(lr){6-7} \cmidrule(lr){8-9} \cmidrule(lr){10-11} \cmidrule(lr){12-13} \cmidrule(lr){14-15} \cmidrule(lr){16-17} \cmidrule(lr){18-19} \cmidrule(lr){20-21}
 $\mathcal{H}(\cdot)$ & $\mathcal{L}_{\text{center}}$ & $\mathcal{L}_{\text{smooth}}$ & $\mathcal{L}_{\text{shuffle}}$ & AA & AA & AF & AA & AF & AA & AF & AA & AF & AA & AF & AA & AF & AA & AF & AA & AF \\
\midrule
$\times$ & $\times$ & $\times$ & $\times$ & \textbf{98.50} & \underline{98.75} & 0.50 & 97.00 & 1.05 & 96.28 & 1.67 & 91.70 & 6.62 & 88.75 & 8.42 & 87.60 & 8.83 & 87.34 & 9.08 & 86.03 & 10.45 \\
$\checkmark$ & $\times$ & $\times$ & $\times$ & \underline{98.00} & \underline{98.75} & \underline{0.00} & 97.50 & \textbf{0.00} & \underline{97.00} & 0.83 & 94.10 & 3.88 & 91.17 & \underline{6.40} & 90.08 & 6.25 & 89.95 & \underline{6.30} & 88.92 & 7.62 \\
$\checkmark$ & $\checkmark$ & $\times$ & $\times$ & \textbf{98.50} & \textbf{99.00} & 0.50 & 97.50 & 1.50 & 96.37 & 1.50 & 94.10 & 3.78 & 91.42 & 6.36 & 91.09 & 6.25 & 90.13 & 6.46 & 89.79 & 7.02 \\
$\checkmark$ & $\checkmark$ & $\checkmark$ & $\times$ & \textbf{98.50} & \textbf{99.00} & \underline{0.00} & 97.00 & 1.00 & \underline{97.00} & \underline{0.67} & \underline{94.30} & \textbf{3.70} & \textbf{92.50} & \textbf{5.26} & \underline{91.57} & \textbf{5.38} & \underline{90.70} & \textbf{5.64} & 90.60 & \textbf{6.21} \\
$\checkmark$ & $\checkmark$ & $\times$ & $\checkmark$ & \textbf{98.50} & \textbf{99.00} & \underline{0.00} & \underline{98.17} & 0.75 & \underline{97.00} & 1.50 & 93.10 & 5.87 & \underline{92.17} & \underline{6.30} & 90.91 & 7.08 & 90.48 & 7.07 & \underline{90.93} & 6.62 \\
\rowcolor{softblue} 
$\checkmark$ & $\checkmark$ & $\checkmark$ & $\checkmark$ & 97.50 & \textbf{99.00} & \textbf{$-$0.50} & \textbf{98.33} & \underline{0.25} & \textbf{97.38} & \textbf{0.50} & \textbf{94.40} & \underline{3.75} & 91.67 & \underline{6.30} & \textbf{91.93} & \underline{5.92} & \textbf{90.84} & \underline{6.36} & \textbf{91.07} & \underline{6.32} \\
\bottomrule
\end{tabular}%
}
\end{table*}

\begin{table}[t] 
\centering
\small
\setlength{\aboverulesep}{0pt}
\setlength{\belowrulesep}{0pt}
\setlength{\tabcolsep}{4pt} 
\renewcommand{\arraystretch}{1.2}
\newcolumntype{Y}{>{\centering\arraybackslash}X}
\setlength{\abovecaptionskip}{7pt}
\caption{Further ablation study of our temporal coherence learning strategy (Sec.~\ref{sec:temporal_learning}, \%). 
\normalfont
``w/o all'' denotes the backbone with the hyperspherical head $\mathcal{H}(\cdot)$ and $\mathcal{L}_{\text{center}}$, our method includes $\mathcal{L}_{\text{smooth}}$ and $\mathcal{L}_{\text{shuffle}}$,``w/ FTCN'' and ``w/ AltFreezing'' adapt designs from ~\cite{zheng2021exploring,wang2023altfreezing}.
Metrics: \textbf{$\text{Acc}_{shuf}$} and \textbf{$\text{Acc}_{swap}$} denote accuracies under segment shuffling and adjacent frame swapping, \textbf{$\text{Acc}_{gen}$} is the generalization accuracy as defined in the last column of Tab.~\ref{tab:generalization_result}.
}
\label{tab:temporal_ablation}

\begin{tabularx}{\columnwidth}{l YYYY}
\toprule
\textbf{Method} & \textbf{mAA} & \textbf{$\text{Acc}_{shuf}$} & \textbf{$\text{Acc}_{swap}$} & \textbf{$\text{Acc}_{gen}$} \\
\midrule
w/o all & 94.21 & 10.50 & 5.80  & 88.40 \\
w/ $\mathcal{L}_{\text{smooth}}$  & \underline{94.57} & 15.20 & 10.94 & \underline{90.65} \\
w/ $\mathcal{L}_{\text{shuffle}}$ & 94.47 & \underline{99.62} & \underline{40.26} & 90.40 \\
\rowcolor{softblue}
\textbf{SphereVideo (ours)} & \textbf{94.68} & \textbf{99.70} & \textbf{50.13} & \textbf{91.15} \\
\midrule
w/ FTCN \cite{zheng2021exploring}            & 88.66 & 22.08 & 31.51 & 75.85 \\
w/ AltFreezing \cite{wang2023altfreezing}    & 93.35 & 11.96 & 12.00 & 87.90 \\
\bottomrule
\end{tabularx}
\end{table}
\subsection{Ablation Analysis}
Tab.~\ref{tab:ablation} and Tab.~\ref{tab:temporal_ablation} present the ablation studies of our proposed method, demonstrating the effectiveness of each individual module.

\noindent\textbf{Impact of Prototype-anchored Hyperspherical Boundary and Centripetal Loss.}
As shown in Tab.~\ref{tab:ablation}, the baseline achieves the lowest Average Accuracy (AA) and the highest Average Forgetting (AF). This suggests that traditional hyperplane boundaries struggle to accommodate diverse and evolving artifacts. By replacing the hyperplane with a hyperspherical head, we observe a significant increase in AA and a notable decrease in AF. This confirms that the proposed hyperspherical boundary, which explicitly defines the real region, better aligns with the task characteristics: real features are clustered within a compact region, whereas diverse fake samples are distributed away from it. Furthermore, the introduction of our centripetal loss constrains real features into an even more compact region, leading to a further improvement in AA.

\noindent\textbf{Effectiveness of Temporal Coherence Learning Strategy.} 
Integrating temporal losses yield marginal AA gains and stable AF under Protocol \ref{prot:Protocol1} in Tab.~\ref{tab:ablation}. This is primarily because current video generation models leave noticeable spatial artifacts, allowing detectors to achieve high accuracy by relying solely on spatial cues for seen data. 
To highlight the efficacy of our temporal learning strategy, Tab.~\ref{tab:temporal_ablation} summarizes specialized ablation studies.  
Following AltFreezing~\cite{wang2023altfreezing}, we generate purely temporal artifacts by reordering real data under two challenging configurations: 1) segment shuffling (consistent with our method) and 2) adjacent frames swapping. The latter is particularly difficult as subtle inter-frame variations require fine-grained temporal modeling. We define $\text{Acc}_{shuf}$ and $\text{Acc}_{swap}$ to measure the accuracy of classifying such data as non-real.
A comparison between first four rows of Tab.~\ref{tab:temporal_ablation} reveals that our method yields substantial improvements and more effectively captures temporal inconsistencies, with gains of 89.20\% ($\text{Acc}_{shuf}$) and 44.33\% ($\text{Acc}_{swap}$). Furthermore, the combination of the two losses proves mutually reinforcing, yielding superior results compared to using either loss alone. Notably, improved OOD generalization ($\text{Acc}_{gen}$) support our idea that temporal consistency learning enables the model to better capture the intrinsic characteristics of real data and form a more discriminative decision boundary. 

\noindent\textbf{Comparison with Other Temporal Strategies.} 
As shown in Tab.~\ref{tab:temporal_ablation}, integrating FTCN~\cite{zheng2021exploring} and AltFreezing~\cite{wang2023altfreezing} into our framework leads to performance drops in both mAA and $\text{Acc}_{gen}$ even when compared to ``w/o all''. 
This indicates that restricting spatial kernels or adopting alternating training is not well aligned with our framework and may disrupt feature learning, hindering the learning of a representative real prototype and a stable boundary. 
Moreover, they are significantly weaker than our SphereVideo framework in capturing temporal incoherence, further demonstrating the effectiveness of our temporal modeling strategy.
\subsection{Robustness Evaluation}
\label{sec:Robustness}
\begin{figure}[t]  
  \centering
  \includegraphics[width=\linewidth]{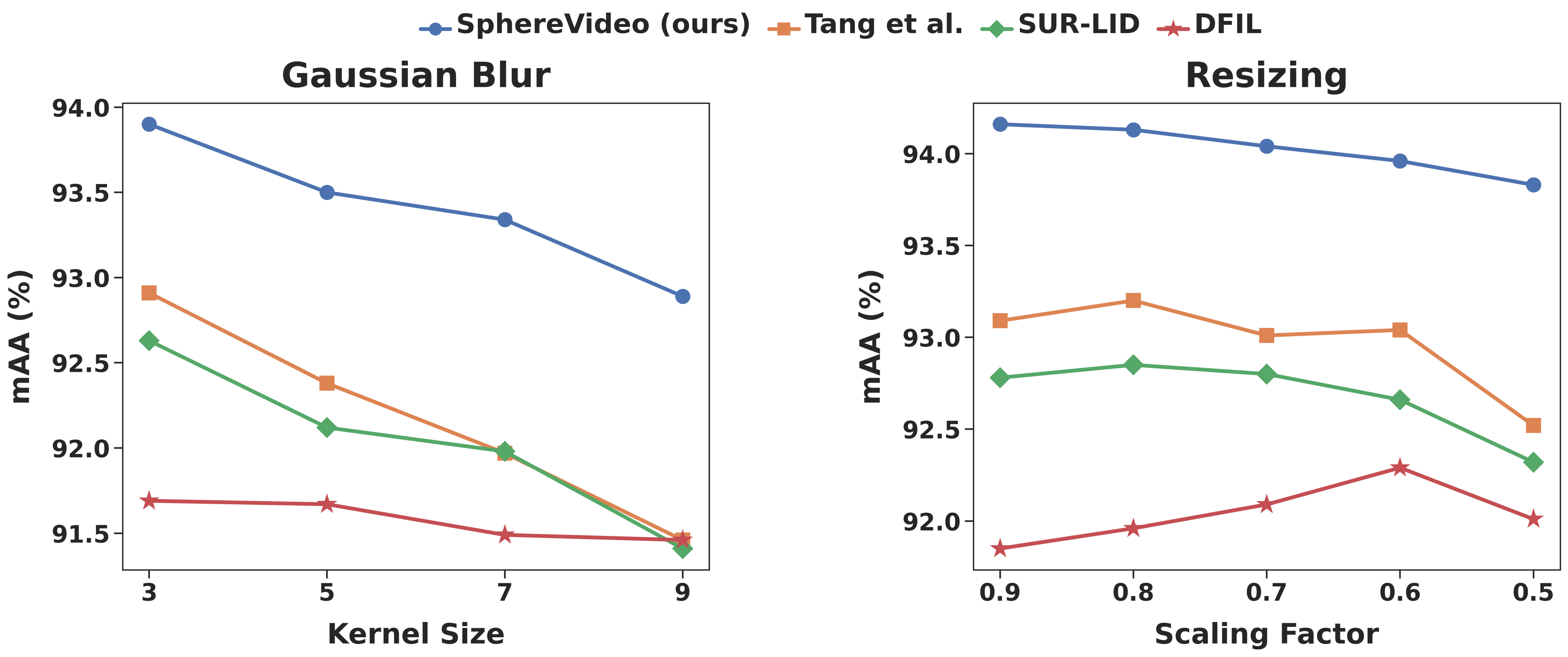} 
  \setlength{\abovecaptionskip}{-3pt}
  \setlength{\belowcaptionskip}{-5pt}
  \caption{Robustness evaluation of different methods.
  \normalfont We compare our method with the four strongest baselines from Tab.~\ref{tab:CL-result} to better visualize performance degradation under varying intensities.}
  \Description{Robustness Evaluation of our framework.}
  \label{fig:robustness}
\end{figure}

We apply data augmentation following DeepfakeBench~\cite{yan2023deepfakebench} during training, while our testing covers both seen and unseen corruptions, focusing on Gaussian blur and resolution resizing to ensure a comprehensive assessment. 
Blur kernel sizes include $\{3 \times 3, 5 \times 5, 7 \times 7, 9 \times 9\}$, and resizing employs scaling factors of $\{0.5, 0.6, 0.7, 0.8, 0.9\}$. Specifically, resizing consists of downsampling followed by upsampling. 
As shown in Fig.~\ref{fig:robustness}, our method achieves the highest mAA across all levels and degrades gracefully, demonstrating superior robustness. 
This strongly supports our hypothesis: isolating feature direction from magnitude forces the model to capture highly robust and invariant representations.

\subsection{In-depth Visualization and Analysis}
\label{Sec:Visualization}
\begin{figure}[t]  
  \centering
  \includegraphics[width=\linewidth]{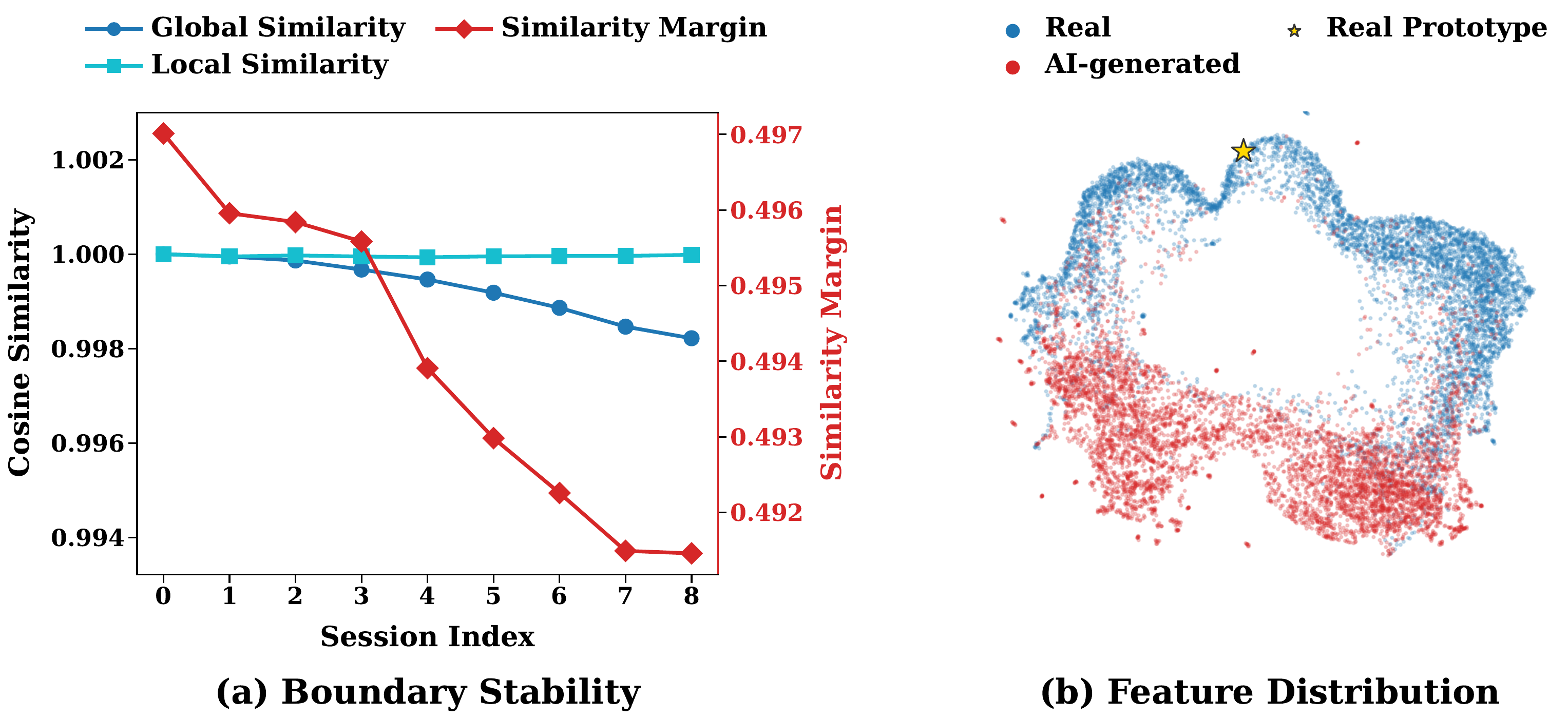} 
\setlength{\abovecaptionskip}{-3pt}
  \setlength{\belowcaptionskip}{-10pt}
  \caption{In-depth visualization. 
  \normalfont
  (a) Global and local similarities measure the cosine similarity of the current prototype $\mathbf{p}_{r}$ against the initial (Session 0) and previous sessions, respectively. The similarity margin $m$ is dynamically updated across sessions. (b) UMAP visualization of the feature distribution at the final session.}
  \Description{Semantic Stability of our framework. }
  \label{fig:in_depth_visual}
\end{figure}

\noindent\textbf{Analysis of Boundary Stability.} The stability of our boundary is illustrated in Fig.~\ref{fig:in_depth_visual}a. Notably, the local similarity of the real prototype remains close to $1.0$, and the global similarity stays above $0.998$ despite exposure to diverse generative artifacts across nine sessions. Meanwhile, the margin $m$ varies slightly from $0.497$ to approximately $0.492$. 
These observations indicate that our model learns a stable prototype representing the intrinsic characteristics of real data while maintaining a well-controlled margin, thereby establishing a robust decision boundary.

\noindent\textbf{Visualization of Feature Distribution.}
Given that we adopt a hyperspherical head, we use UMAP to visualize feature distributions, as it preserves the global manifold structure during dimensionality reduction. As shown in Fig.~\ref{fig:in_depth_visual}b, the projected features form a ring-like structure. Real features exhibit a clear centripetal distribution in the upper arc around the learned real prototype $\mathbf{p}_{r}$, while fake samples lie in the lower segment of the ring. This clear separation suggests the effectiveness of our SphereVideo framework in distinguishing AI-generated videos from real ones.

\section{Conclusion}
We propose SphereVideo framework for continual AI-generated video detection. Motivated by the compact feature distribution of real videos, we introduce a hyperspherical decision boundary that defines a compact real region centered around a real prototype. The prototype serves as a stable anchor for continual learning, thereby mitigating catastrophic forgetting. Furthermore, to mitigate shortcut learning on spatial artifacts, we propose a strategy that imposes constraints on real data across different feature levels, thereby enhancing temporal modeling while facilitating a more representative real prototype and a more stable boundary.
Extensive experiments on our comprehensive benchmark under both continual learning and generalization protocols demonstrate the effectiveness of SphereVideo.


\begin{acks}
This work was supported by NSFC projects (No. 62232006, No. 62522206, and No. 62521004).
\end{acks}

\bibliographystyle{ACM-Reference-Format}
\bibliography{main}
\clearpage
\appendix
\section{Overview}
This supplementary material is organized as follows:
\begin{itemize}[nosep, leftmargin=*] 
\item Dataset Introduction
\item Additional Experiments under Protocol 1
\item Additional Experiments under Protocol 2
\item Detailed Performance of Each Method under Protocol 1
\item Additional Boundary Stablility Analysis
\item Grad-CAM Visualization
\item Different Dataset Orders
\item Source of Fig.~\ref{fig:motivation}a
\item High-Quality Video Evaluation
\item Additional Robustness Evaluations
\end{itemize}

\section{Dataset Introduction}
To systematically evaluate the performance of AI-generated video (AIGV) detectors under a continual learning (CL) paradigm, we construct a comprehensive and challenging benchmark. As detailed in Sec.~\ref{Sec:ExperimentalSetting}, our benchmark encompasses three primary generative paradigms: UNet-based diffusion, DiT-based diffusion, and autoregressive models~\cite{ma2025controllable}. We further incorporate real-world videos from multiple sources to mitigate potential dataset biases, covering a wide range of content, resolutions, frame rates, and durations.

For each continual training session, the dataset consists of paired real-world and AI-generated videos. By aligning diverse generative data from a wide range of generative models with high-quality real-world counterparts, our benchmark effectively reflects the complexity of AIGV detection in practical scenarios. Detailed statistics of these paired datasets are summarized by generative paradigm in Tab.~\ref{tab:training_dataset_final}, including model names, source datasets, and key attributes such as quantity, frame rate, resolution, and duration.

Notably, the benchmark spans a wide range of spatial resolutions (480p to 1280p), temporal rates (8 to 30 FPS), and video durations (from 1 seconds to one hour), introducing substantial domain gaps that rigorously evaluate detector robustness. All datasets are split into training and testing sets with a 9:1 ratio using a fixed random seed of 100 to ensure reproducibility. Furthermore, as described in the Implementation Details, we sample a fixed number of frames from each video during both training and testing, thereby mitigating biases related to video duration.

Additionally, the open-world benchmark for generalization evaluation is adopted from GenBuster-200K~\cite{wen2025busterx}; further details can be found in the original work.
\begin{table*}[t]
\centering
\setlength{\abovecaptionskip}{7pt}
\caption{Details of the dataset in protocol 1.
\normalfont
We pair AI-generated videos with their corresponding real-world counterparts, forming the training and evaluation data for each session.
“Qty.” denotes the number of videos, and “Dur.” denotes the duration.}
\label{tab:training_dataset_final}
\resizebox{\textwidth}{!}{%
\begin{tabular}{@{}ll rcc cc l rccc@{}}
\toprule
\multicolumn{7}{c}{\textbf{AI-generated Videos}} & \multicolumn{5}{c}{\textbf{Real Videos}} \\
\cmidrule(r){1-7} \cmidrule(l){8-12}
\textbf{Category} & \textbf{Generative Method} & \textbf{Qty.} & \textbf{FPS} & \textbf{Resolution} & \textbf{Source} & \textbf{Dur.} & \textbf{Dataset} & \textbf{Qty.} & \textbf{FPS} & \textbf{Resolution} & \textbf{Dur.} \\
\midrule
\multirow{3}{*}{UNet-based}
& SVD~\cite{blattmann2023stable} & 1,000 & 8 & 1024$\times$576 & I2V & 4s & YouTube-8M~\cite{abu2016youtube} & 1,000 & 30 & 1280$\times$720 & 4--5s \\
\cmidrule(lr){2-12}
& Zeroscope & 800 & \multirow{2}{*}{8} & \multirow{2}{*}{1024$\times$576} & \multirow{2}{*}{T2V} & 4s & \multirow{2}{*}{InternVid-10M~\cite{wang2023internvid}} & 800 & \multirow{2}{*}{30} & \multirow{2}{*}{1280$\times$720} &  \multirow{2}{*}{1--30s} \\
& VideoCrafter1~\cite{chen2023videocrafter1} & 450 & & & & 2s & &  450& & & \\
\midrule
\multirow{4}{*}{DiT-based}
& HunyuanVideo~\cite{kong2024hunyuanvideo} & \multirow{4}{*}{1,000} & \multirow{4}{*}{24} & \multirow{4}{*}{1024$\times$1024} & \multirow{4}{*}{T2V} & \multirow{4}{*}{5s} & \multirow{4}{*}{OpenVid-1M~\cite{nan2024openvid}} & \multirow{4}{*}{1,000} & \multirow{4}{*}{24} & \multirow{4}{*}{1024$\times$1024} & \multirow{4}{*}{5s} \\
& CogVideoX~\cite{yang2024cogvideox} & & & & & & & & & & \\
& LTX-Video~\cite{hacohen2024ltx} & & & & & & & & & & \\
& EasyAnimateV5.1~\cite{xu2024easyanimate} & & & & & & & & & & \\
\midrule
\multirow{2}{*}{Autoregressive}
& Magi-1~\cite{teng2025magi} & \multirow{2}{*}{1000} & 24 & 720$\times$720 & \multirow{2}{*}{T2V} & 4s & \multirow{2}{*}{HD-VG-130M~\cite{wang2023videofactory}} & \multirow{2}{*}{1,000} & \multirow{2}{*}{30} & \multirow{2}{*}{1280$\times$720} & \multirow{2}{*}{1s--1h} \\
& CausVid~\cite{yin2025slow} & & 16 & 832$\times$480 & & 5s & & & & & \\
\bottomrule
\end{tabular}
}
\end{table*}

\section{Additional Experiments under Protocol 1}
\label{sec:Protocol1}
We evaluate our method against several state-of-the-art non-replay continual learning baselines~\cite{sun2025continual,zhou2025dual} under the sequential learning setting defined in Protocol~1. The comprehensive results, presented in Tab.~\ref{tab:CL-resultv2}, demonstrate the clear superiority of our method.

Our method consistently outperforms all baselines, achieving the highest Average Accuracy (AA) and the lowest Average Forgetting (AF) across nearly all sessions. The performance gap becomes particularly pronounced from the \textit{Magi-1} session onwards. This can be attributed to the design of the benchmark: the initial four sessions involve models from the same generative paradigm (DiT-based diffusion), which tend to share similar artifact patterns, making the early sessions relatively less challenging. As a result, all methods perform comparably early on. However, as the sequence progresses to generative models from fundamentally different paradigms (e.g., \textit{Magi-1}), the distribution shift becomes significantly larger. In this more challenging and diverse setting, the advantages of our method in robustness and adaptability become more evident.

Furthermore, the superior performance on new data (New.Acc) indicates that our model has strong adaptability to new data; combined with the low AF, this suggests that our method achieves a favorable stability-plasticity trade-off.

Nevertheless, a performance gap persists between non-replay methods, including ours, and replay-based methods. This gap underscores the challenges of continual learning without memory buffers in video-level tasks, as the complex spatio-temporal dynamics inherent in video data introduce substantial distribution variability, while the high-capacity video encoders required to jointly model spatial and temporal information are more prone to representation drift, making it difficult to maintain stable representations. This, in turn, highlights an important direction for future research.
 
\begin{table*}[t]
\centering
\small
\setlength{\aboverulesep}{0pt}
\setlength{\belowrulesep}{0pt}
\setlength{\tabcolsep}{2pt}
\renewcommand{\arraystretch}{1.2}
\setlength{\abovecaptionskip}{7pt}
\caption{Comparison with non-replay continual learning methods across 9 sessions under protocol 1 (\%).
\normalfont
For a fair comparison, our method (highlighted in blue) is also replay-free. Numerical prefixes (e.g., "0-") denote the session index. 
Bold and underlined values represent the best and second-best results, respectively.}
\label{tab:CL-resultv2}

\begin{tabularx}{\dimexpr\textwidth-5pt\relax}{ l ccc c H c SS c SS c SS }
\toprule
\multirow{2}{*}{\textbf{Method}} 
& \multirow{2}{*}{\makecell{\textbf{Conference}\\\textbf{\& Year}}}
& \multirow{2}{*}{\makecell{\textbf{Continual}\\\textbf{Learning}}}
& \multirow{2}{*}{\makecell{\textbf{Replay}\\\textbf{Set}}}
& & \textbf{0-HunyuanVideo}
& & \multicolumn{2}{c}{\textbf{1-EasyAnimate}}
& & \multicolumn{2}{c}{\textbf{2-CogVideoX}}
& & \multicolumn{2}{c}{\textbf{3-LTX-Video}} \\
\cmidrule(lr){6-6} \cmidrule(lr){8-9} \cmidrule(lr){11-12} \cmidrule(lr){14-15}
& & & & & AA & & AA & AF & & AA & AF & & AA & AF \\
\midrule
HDP~\cite{sun2025continual} & IJCV 2024 & $\checkmark$ & $\times$ & & \textbf{99.00} & & \textbf{98.25} & 2.50 & & 96.67 & 2.75 & & \underline{95.25} & 3.67 \\
DUCT~\cite{zhou2025dual} & CVPR 2025 & $\checkmark$ & $\times$ & & \underline{98.00} & & \underline{98.00} & \underline{1.50} & & \underline{97.00} & \underline{2.00} & & 95.12 & \underline{3.17} \\
\rowcolor{softblue}
\textbf{SphereVideo (ours)} & - & $\checkmark$ & $\times$ & & \underline{98.00} & & \textbf{98.25} & \textbf{$-$0.50} & & \textbf{98.50} & \textbf{$-$0.25} & & \textbf{96.12} & \textbf{2.00} \\
\bottomrule
\end{tabularx}

\vspace{3mm}

\begin{tabularx}{\dimexpr\textwidth-5pt\relax}{ l c YY c YY c YY c YY c YY c Y }
\toprule
\multirow{2}{*}{\textbf{Method}} 
& & \multicolumn{2}{c}{\textbf{4-Magi-1}}
& & \multicolumn{2}{c}{\textbf{5-CausVid}}
& & \multicolumn{2}{c}{\textbf{6-ZeroScope}}
& & \multicolumn{2}{c}{\textbf{7-VideoCrafter1}}
& & \multicolumn{2}{c}{\textbf{8-SVD}}
& & \multirow{2}{*}{\makecell{\textbf{New.}\\\textbf{Acc}}} \\
\cmidrule(lr){3-4} \cmidrule(lr){6-7} \cmidrule(lr){9-10} \cmidrule(lr){12-13} \cmidrule(lr){15-16}
& & AA & AF & & AA & AF & & AA & AF & & AA & AF & & AA & AF & & \\
\midrule
HDP~\cite{sun2025continual}  & & 80.60 & 21.00 & & \underline{76.75} & \underline{24.40} & & \underline{58.45} & \underline{45.25} & & \textbf{76.64} & \underline{24.95} & & \underline{64.53} & \underline{36.24} & & \underline{96.10} \\
DUCT~\cite{zhou2025dual}  & & \underline{82.90} & \underline{17.62} & & 76.42 & \underline{24.40} & & 57.43 & 46.00 & & 69.67 & 30.57 & & 63.75 & 36.88 & & 95.35 \\
\rowcolor{softblue}
\textbf{SphereVideo (ours)} & & \textbf{93.90} & \textbf{4.25} & & \textbf{87.00} & \textbf{11.60} & & \textbf{67.75} & \textbf{34.08} & & \underline{74.91} & \textbf{24.84} & & \textbf{74.15} & \textbf{25.61} & & \textbf{96.54} \\
\bottomrule
\end{tabularx}
\end{table*}

\section{Additional Experiments under Protocol 2}
We evaluate our non-replay method against several state-of-the-art non-replay continual learning baselines under the open-world benchmark defined in Protocol 2. The comprehensive results in Tab.~\ref{tab:generalization_resultv2} demonstrate the clear superiority of our method.

Our method achieves the best or second-best performance across all metrics. Compared with HDP~\cite{sun2025continual}, although it attains slightly higher accuracy on certain unseen generative models, its performance on paired real data is significantly lower, leading to a clearly inferior overall average.
Nevertheless, as discussed in Sec.~\ref{sec:Protocol1}, video-level tasks are inherently challenging, and non-replay methods are generally less stable during training than replay-based methods. Consequently, the learned data distributions are less well preserved, resulting in inferior generalization performance. 
\begin{table*}[t]
\centering
\small
\setlength{\aboverulesep}{0pt}
\setlength{\belowrulesep}{0pt}
\setlength{\tabcolsep}{2.5pt}
\renewcommand{\arraystretch}{1.2}
\setlength{\abovecaptionskip}{7pt}
\caption{Comparison with non-replay continual learning methods under protocol 2 (\%).
\normalfont
Our method is also replay-free. 
Following the setting of the benchmark~\cite{wen2025busterx}, we report accuracy across OOD generative models, along with accuracy on paired real samples (Real.Acc), overall accuracy on all OOD fake samples (Fake.Acc), and their mean (Avg.Acc).}
\label{tab:generalization_resultv2}

\begin{tabularx}{\dimexpr\textwidth-5pt\relax}{ l c YYYYYYYY c YYY }
\toprule
\multirow{2}{*}{\textbf{Method}} 
& & \multicolumn{8}{c}{\textbf{Out-of-Distribution (OOD) Generative Models (Acc)}} 
& & \multirow{2}{*}{\makecell{\textbf{Real.}\\\textbf{Acc}}} 
& \multirow{2}{*}{\makecell{\textbf{Fake.}\\\textbf{Acc}}} 
& \multirow{2}{*}{\makecell{\textbf{Avg.}\\\textbf{Acc}}} \\
\cmidrule(lr){3-10}
& & \textbf{Gen3} & \textbf{Jimeng} & \textbf{Kling} & \textbf{Luma} & \textbf{Pika} & \textbf{Sora} & \textbf{Vidu} & \textbf{Wanx} & & & & \\
\midrule
HDP~\cite{sun2025continual} & & \textbf{75.00} & \textbf{60.00} & \underline{61.00} & \textbf{79.00} & \textbf{87.00} & \underline{71.00} & 92.67 & 80.67 & & 66.90 & \textbf{76.40} & \underline{71.65} \\
DUCT~\cite{zhou2025dual}  & & 61.00 & 28.00 & 45.00 & 50.00 & 83.00 & 53.50 & 82.67 & 66.67 & & \underline{76.10} & 59.80 & 67.95 \\
\rowcolor{softblue} 
\textbf{SphereVideo (ours)} & & \underline{72.00} & \underline{47.00} & \textbf{78.00} & \underline{71.00} & \underline{81.00} & \textbf{73.00} & \textbf{94.67} & \textbf{81.33} & & \textbf{78.30} & \underline{75.90} & \textbf{77.10} \\
\bottomrule
\end{tabularx}
\end{table*}

\section{Detailed Performance of Each Method under Protocol 1}
\begin{figure*}[t] 
  \centering
  \includegraphics[width=1\textwidth]{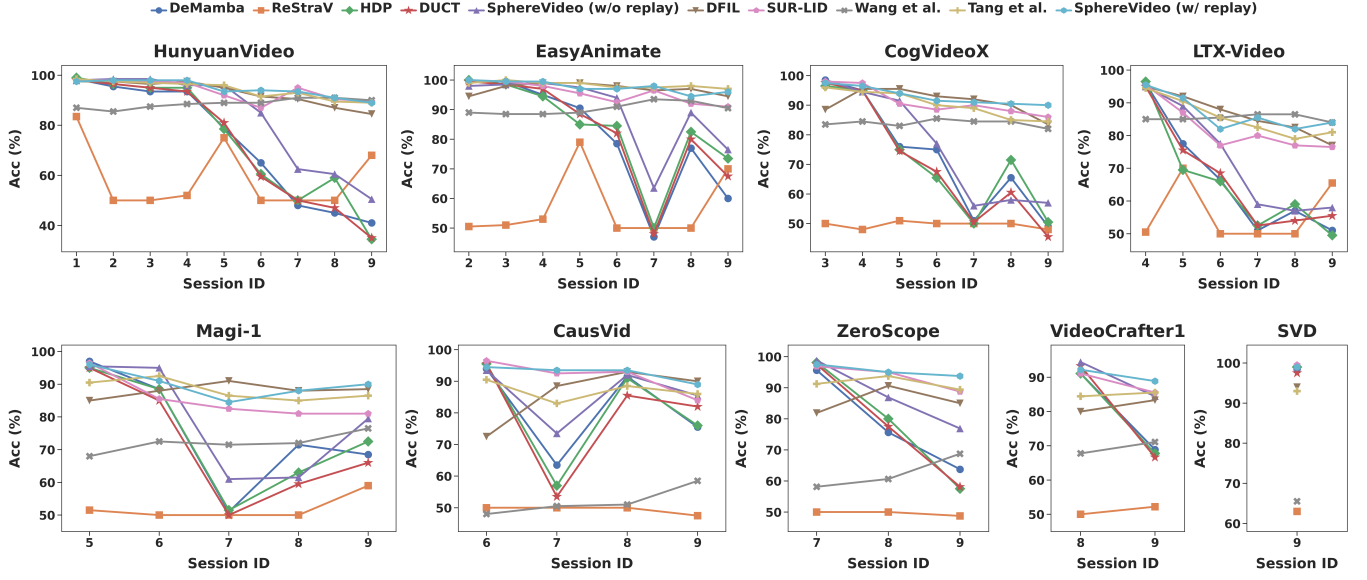} 
  \setlength{\abovecaptionskip}{-3pt}
  \setlength{\belowcaptionskip}{-3pt} 
  \caption{Detailed performance of each method under protocol 1. \normalfont
  The line charts show the performance on each dataset under protocol 1, where the lines show the performance variation of different methods on a particular dataset across different sessions.}
  \label{fig:CL_Accuracy_Curves}
\end{figure*}
As shown in Fig.~\ref{fig:CL_Accuracy_Curves}, we visualize the detailed performance of each method across sessions under protocol 1 using line charts. Our method (SphereVideo w/ replay) demonstrates a clear advantage, maintaining competitive performance throughout most of the training sessions. Although the replay-free version (SphereVideo w/o replay) exhibits more pronounced forgetting, it still outperforms other non-replay methods, highlighting its effectiveness even without memory buffers.

\section{Additional Boundary Stablility Analysis}
\begin{figure}[t]  
  \centering
  \includegraphics[width=0.9\linewidth]{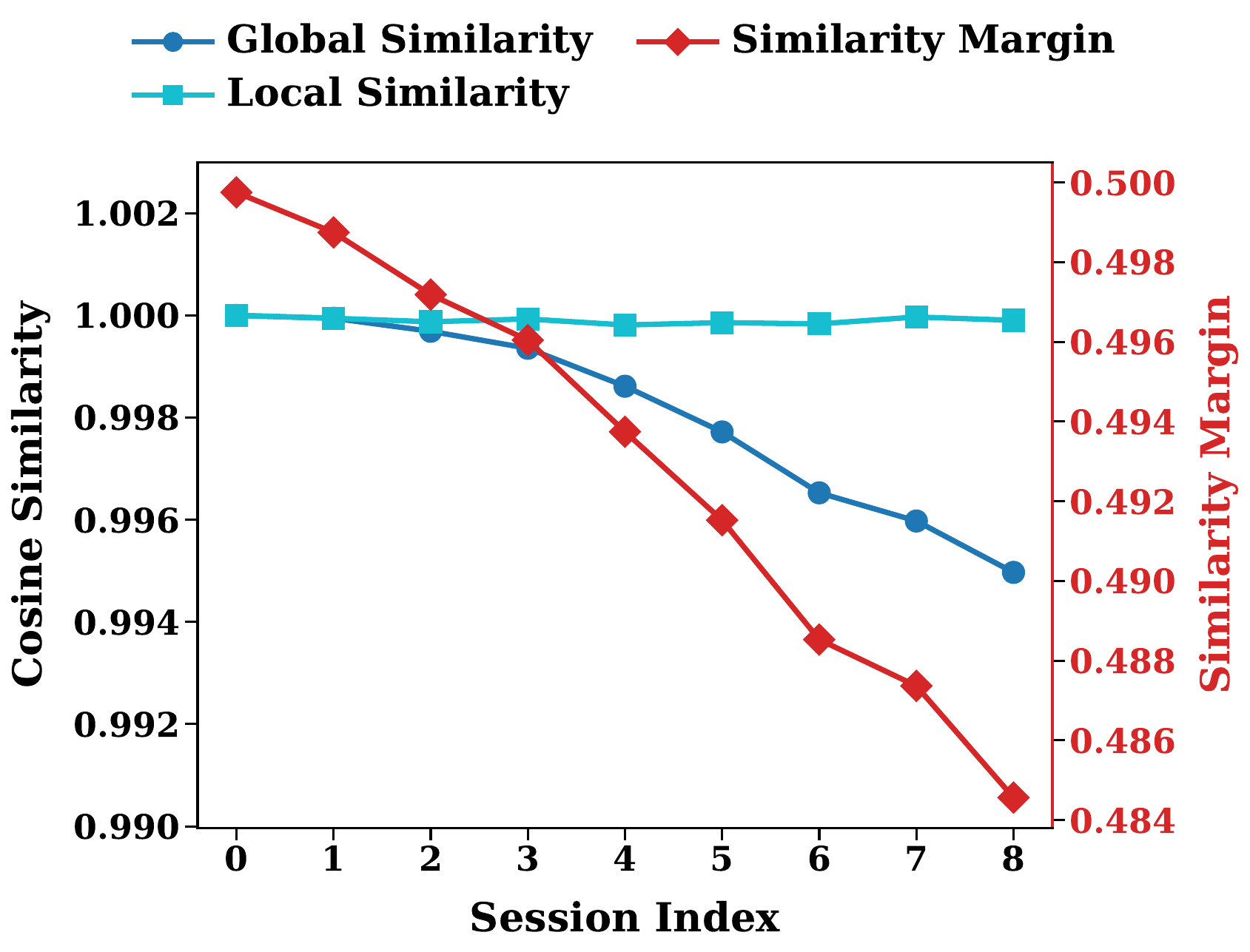} 
    \setlength{\belowcaptionskip}{-3pt} 
    \caption{Visualization of boundary stability for our method (using only the hyperspherical head).
    \normalfont
    Global and local similarities measure the cosine similarity of the current prototype $\mathbf{p}_{r}$ against the initial (Session 0) and previous sessions, respectively. The similarity margin $m$ is dynamically updated across sessions.
    }
    \label{fig:SphereVideo_Qualitative_Analysis_Clean_baseline}
\end{figure}
\begin{figure}[t]  
  \centering
  \includegraphics[width=0.9\linewidth]{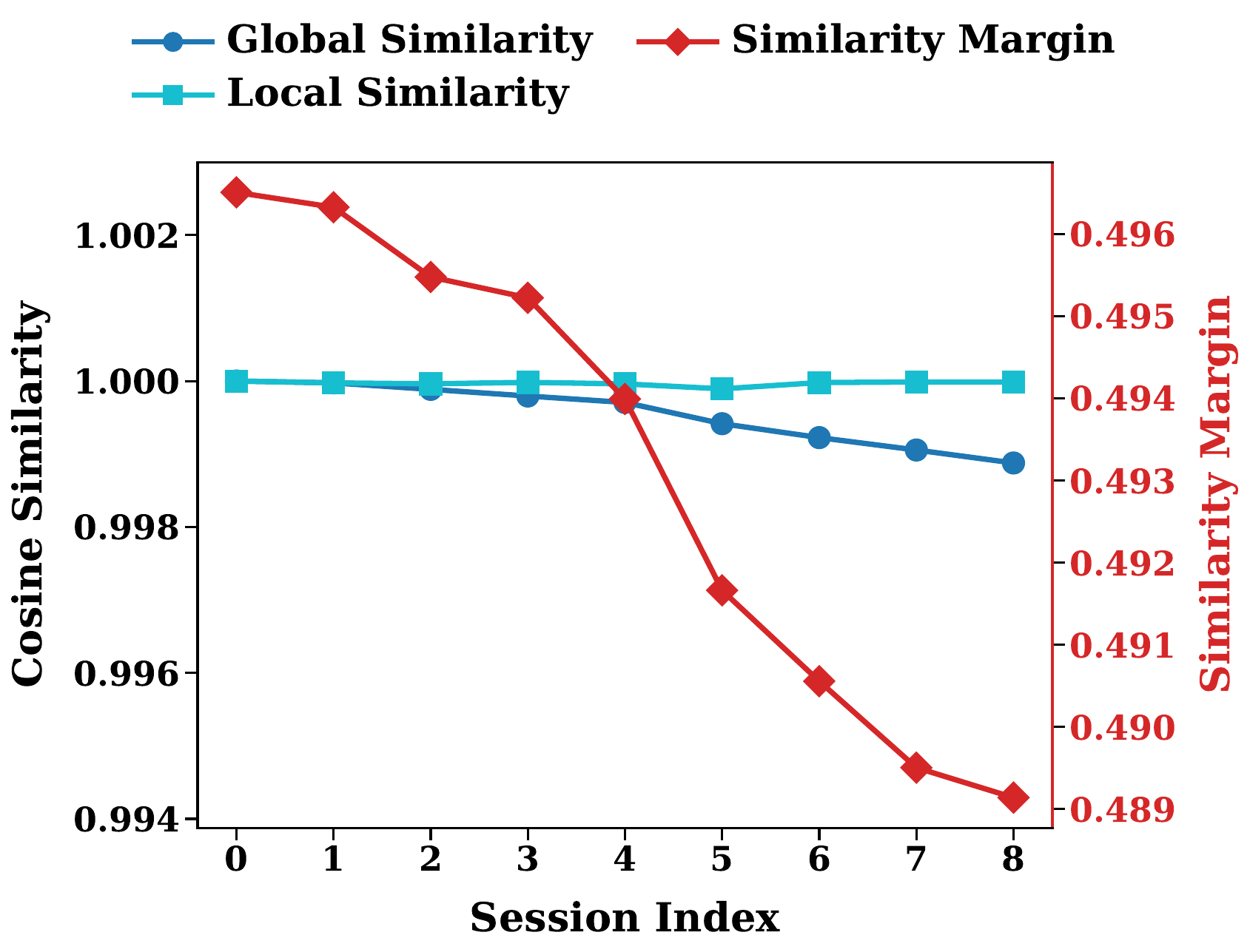} 
    \setlength{\belowcaptionskip}{-3pt} 
   \caption{Visualization of boundary stability for our method (using only the hyperspherical head with centripetal loss).}
\label{fig:SphereVideo_Qualitative_Analysis_Clean_baselinev2}
\end{figure}
As shown in Fig.\ref{fig:SphereVideo_Qualitative_Analysis_Clean_baseline} and Fig.\ref{fig:SphereVideo_Qualitative_Analysis_Clean_baselinev2}, we visualize the boundary stability using solely the hyperspherical head and incorporating the centripetal loss, respectively. In contrast, Fig.~\ref{fig:in_depth_visual} in the main text presents the final results after further integrating the temporal coherence learning strategy.

Due to the diverse datasets encountered during continual learning, real data come from different sources and include a wide range of content, resolutions, and frame rates. Consequently, a slight expansion of the real region during training is normal. This phenomenon likely arises because the initial real datasets are limited in size, and as increasingly diverse datasets are introduced, the boundary needs to expand slightly to maintain training stability. Notably, our method exhibits only minor changes, with the margin $m$ varying slightly from $0.497$ to approximately $0.492$.

In contrast, as shown in Fig.~\ref{fig:SphereVideo_Qualitative_Analysis_Clean_baseline}, using only the hyperspherical head without additional constraints leads to less stable real prototypes $\mathbf{p}_{r}$ and similarity margins $m$ compared to our full method (presented in Fig.~\ref{fig:in_depth_visual}a). With the addition of the centripetal loss (Fig.~\ref{fig:SphereVideo_Qualitative_Analysis_Clean_baselinev2}), the real prototype becomes much more stable, and the reduction in the similarity margin is mitigated.
Building on this, incorporating the Temporal Coherence Learning Strategy (as described in Sec.~\ref{Sec:Visualization}) further enhances the model’s ability to learn representative real prototypes and compact real regions, thereby minimizing the decrease in the similarity margin even further.

Therefore, compared to methods without any constraints, our full SphereVideo framework, which incorporates both the centripetal loss and the temporal coherence learning constraint, achieves the most stable performance, learning relatively more compact real regions and more representative prototypes.

\section{Grad-CAM Visualization}
\begin{figure*}[t]  
  \centering
  \includegraphics[width=\linewidth]{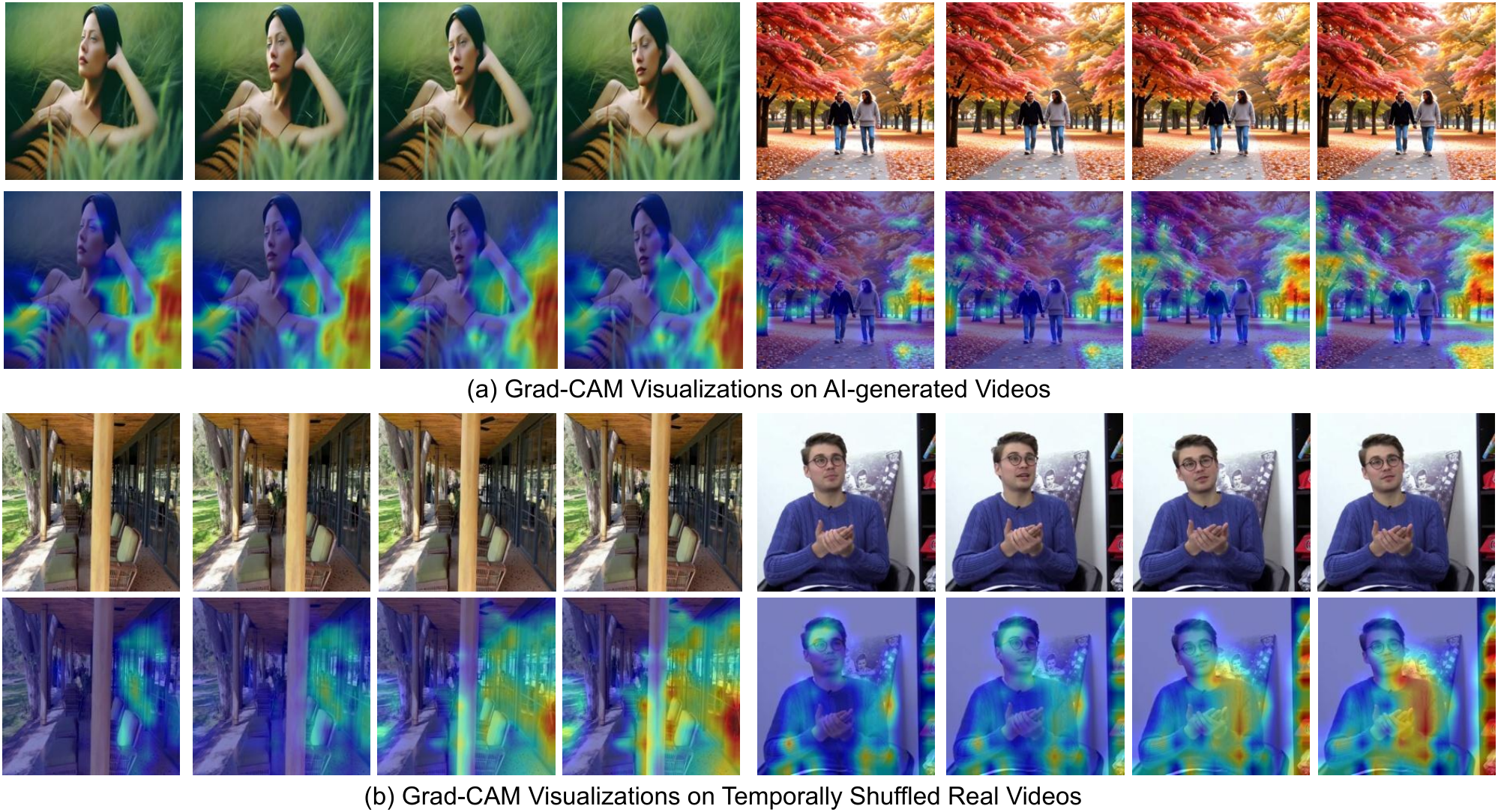} 
    \setlength{\abovecaptionskip}{-3pt}
   \caption{Grad-CAM visualization results of the proposed SphereVideo framework. 
\normalfont
(a) Results on AI-generated videos sampled from our continual learning benchmark. 
(b) Results on real videos after segment shuffling, where each clip is divided into two segments and their temporal order is permuted, following our method design.
}
\label{fig:cam_visualization}
\end{figure*}
In Fig.~\ref{fig:cam_visualization}, we present Grad-CAM results under two types of data. We observe that our method produces clear activation responses for both AI-generated videos and temporally shuffled inputs.

For the examples in Fig.~\ref{fig:cam_visualization}a, SphereVideo focuses on unrealistic regions, such as unnatural grass and trees, resulting in high activation values, thereby highlighting its ability to capture artifacts in AI-generated videos.

For the examples in Fig.~\ref{fig:cam_visualization}b, temporal shuffling disrupts global consistency while preserving local structures. Under this perturbation, the first two frames remain locally coherent and free of artifacts. However, starting from the third frame, temporal inconsistency emerges, leading to noticeable temporal artifacts. A method sensitive to temporal dynamics is expected to assign higher activation to these later frames. 
Our results confirm that the proposed SphereVideo framework effectively captures such temporal inconsistencies, demonstrating its effectiveness.

\section{Different Dataset Orders}
We explore different sequential dataset orders in continual learning setting to validate the robustness of our method. 

Order 1 follows the sequence defined in protocol 1: \{HunyuanVideo, EasyAnimate, CogVideoX, LTX-Video, Magi-1, CausVid, Zeroscope, VideoCrafter1, SVD\}. In our benchmark construction, datasets are organized according to generative paradigms, where the first four, the middle two, and the last three models belong to the same paradigms, respectively. Therefore, when reshuffling the order, we treat models within the same paradigm as a group and permute these groups.
Accordingly, Order 2 is defined as \{HunyuanVideo, EasyAnimate, CogVideoX, LTX-Video, SVD, Zeroscope, VideoCrafter1, CausVid, Magi-1\}. Order 3 is defined as \{SVD, Zeroscope, VideoCrafter1, HunyuanVideo, EasyAnimate, CogVideoX, LTX-Video, CausVid, Magi-1\}. 

In addition, to assess whether interleaving datasets from different paradigms affects performance stability, we construct Order 4, where data from different paradigms are alternated: \{HunyuanVideo, EasyAnimate, CogVideoX, LTX-Video, SVD, CausVid, Magi-1, Zeroscope, VideoCrafter1\}, and Order5: \{HunyuanVideo, EasyAnimate, CausVid, CogVideoX, SVD, Zeroscope, LTX-Video, VideoCrafter1, Magi-1
\}.

The results in Tab.~\ref{tab:order_ablation} show that varying the dataset order yields comparable results, demonstrating the robustness of our method to different data ordering.

\begin{table}[!t] 
\centering
\setlength{\abovecaptionskip}{7pt}
\caption{Comparison of different sequential dataset orders in continual learning.
\normalfont
\textbf{$\text{Acc}_{gen}$} denotes the generalization accuracy as defined in the last column of Tab.~\ref{tab:generalization_resultv2}.}
\label{tab:order_ablation}
\begin{tabularx}{\columnwidth}{l >{\centering\arraybackslash}X >{\centering\arraybackslash}X >{\centering\arraybackslash}X}
\toprule
\textbf{Setting} & \textbf{mAA} & \textbf{New.Acc} & \textbf{Acc$_{gen}$} \\ 
\midrule
\rowcolor{softblue}
\textbf{Order 1 (Ours)} & \textbf{94.68} & \textbf{96.64} & \textbf{91.15} \\
Order 2 & \underline{94.63} & 96.50 & 90.40 \\
Order 3 & 94.33 & \underline{96.56} & 90.15 \\
Order 4 & \textbf{94.68} & 95.75 & \underline{90.75} \\
Order 5 & 94.50 & 95.32 & 90.25 \\
\bottomrule
\end{tabularx}
\end{table}

\section{Source of Fig.~\ref{fig:motivation}a}
Fig.~\ref{fig:motivation}a visualizes features extracted from our benchmark's test set using the adapted SUR-LID [7] model (with a VideoMAE backbone) trained on the corresponding training set. Notably, this model does not explicitly encourage the features of real samples to aggregate; instead, it tends to separate real features across different sessions. However, real features still exhibit a compact distribution, demonstrating that the inherent structural consistency of real data naturally emerges even in the absence of explicit clustering constraints, laying a solid foundation for our design. 
Furthermore, ``Dataset A/B/C'' and ``Generator X/Y/Z'' denote representative real datasets and generators from our benchmark, simplified for illustration. Owing to the high diversity of data sources and generators in our benchmark, this observation is expected to generalize broadly.

\section{High-Quality Video Evaluation}
Since earlier datasets contain noticeable spatial artifacts, the benefits of the temporal coherence learning strategy (TCLS) are less pronounced. However, experiments in Tab.\ref{tab:temporal_ablation} involving manual temporal perturbations validate the effectiveness of SphereVideo in temporal modeling. To further evaluate temporal artifact modeling under highly realistic fake videos, we curate a high-quality test set comprising 400 videos from SeedDance 2.0 and HappyHorse (200 each), excluding clips with visible frame-level defects.  
Our full model achieves 54.75\% accuracy in direct evaluation (47.50\% w/o TCLS, 40.75\% for SUR-LID ~\cite{cheng2025stacking}, and 33.00\% for Tang et al. ~\cite{tang2025towards}), improving to 93.75\% after fine-tuning.
Overall, SphereVideo effectively detects high-quality AIGVs, especially after fine-tuning, while TCLS further improves the handling of natural temporal artifacts in AIGVs beyond manual perturbations.

\section{Additional Robustness Evaluations}
While Gaussian blur and resizing in the main text (Sec.~\ref{sec:Robustness}) primarily simulate image-level compressions, video-level tasks necessitate temporal-aware compression evaluations. Therefore, we further evaluate our method under video-level H.264 compression, where CRF (Constant Rate Factor) controls the compression trade-off, with higher values indicating stronger compression and lower video quality.

Starting from a baseline mAA of $94.68\%$, under video-level H.264 compression with $\text{CRF} \in \{23, 30, 38, 45\}$, our mAA remains $93.49\%$, $93.28\%$, $93.17\%$, and $93.12\%$, respectively, suffering only a negligible drop. This further demonstrates the strong robustness of our method.
\end{document}